\documentclass[oneeqnum, onethmnum, onefignum, onetabnum]{article}

\usepackage{amssymb, bigints,xcolor, multirow, array, cancel, amsthm,url,subcaption}
\usepackage{graphicx}
\usepackage[top=1.5in, bottom=1.25in, left=1.25in, right=1.25in]{geometry}

\usepackage[ruled]{algorithm2e}

\def\R{\mathbb{R}}
\def\Z{\mathbb{Z}}

\def\P{\mathbf{P}}

\def\f{\mathbf{f}}

\def\prox{\text{prox}}

\graphicspath{{./figures/}}

\title{Restoration of Pansharpened Images by Conditional Filtering in the PCA Domain}

\author{J.~Duran\footnotemark[1] \and A.~Buades\footnotemark[1]}

\begin{document}

\maketitle

\renewcommand{\thefootnote}{\fnsymbol{footnote}}

\footnotetext[1]{Universitat de les Illes Balears, Dept.~Mathematics and Computer Science and IAC3, Cra.~de Valldemossa km.~7.5, E-07122 Palma, Spain (email: \{joan.duran, toni.buades\}@uib.es). These authors were supported by the Ministerio de Econom{\'i}a, Industria y Competitividad of the Spanish Government under grants TIN2014-53772-R and TIN2017-85572-P.}
\renewcommand{\thefootnote}{\arabic{footnote}}

\begin{abstract}
Pansharpening techniques aim at fusing low-resolution multispectral (MS) images and high-resolution panchromatic (PAN) images to produce high-resolution MS images. Despite significant progress in the field, spectral and spatial distortions might still compromise the quality of the results. We introduce a  restoration strategy to mitigate artifacts of fused products. After applying the Principal Component Analysis (PCA) transform to a pansharpened image, the chromatic components are filtered conditionally to the geometry of PAN. The structural component is then replaced by the locally histogram-matched PAN for spatial enhancement. Experimental results illustrate the efficiency of the proposed restoration chain.
\end{abstract}

\pagestyle{myheadings}
\thispagestyle{plain}
\section{Introduction}

\begin{figure}[t!]
\centering
\footnotesize
\begin{tabular}{c@{\hskip 0.02in}c}
\includegraphics[trim= 11.9cm 43.5cm 55.3cm 23.7cm, clip=true, width=0.3\columnwidth]{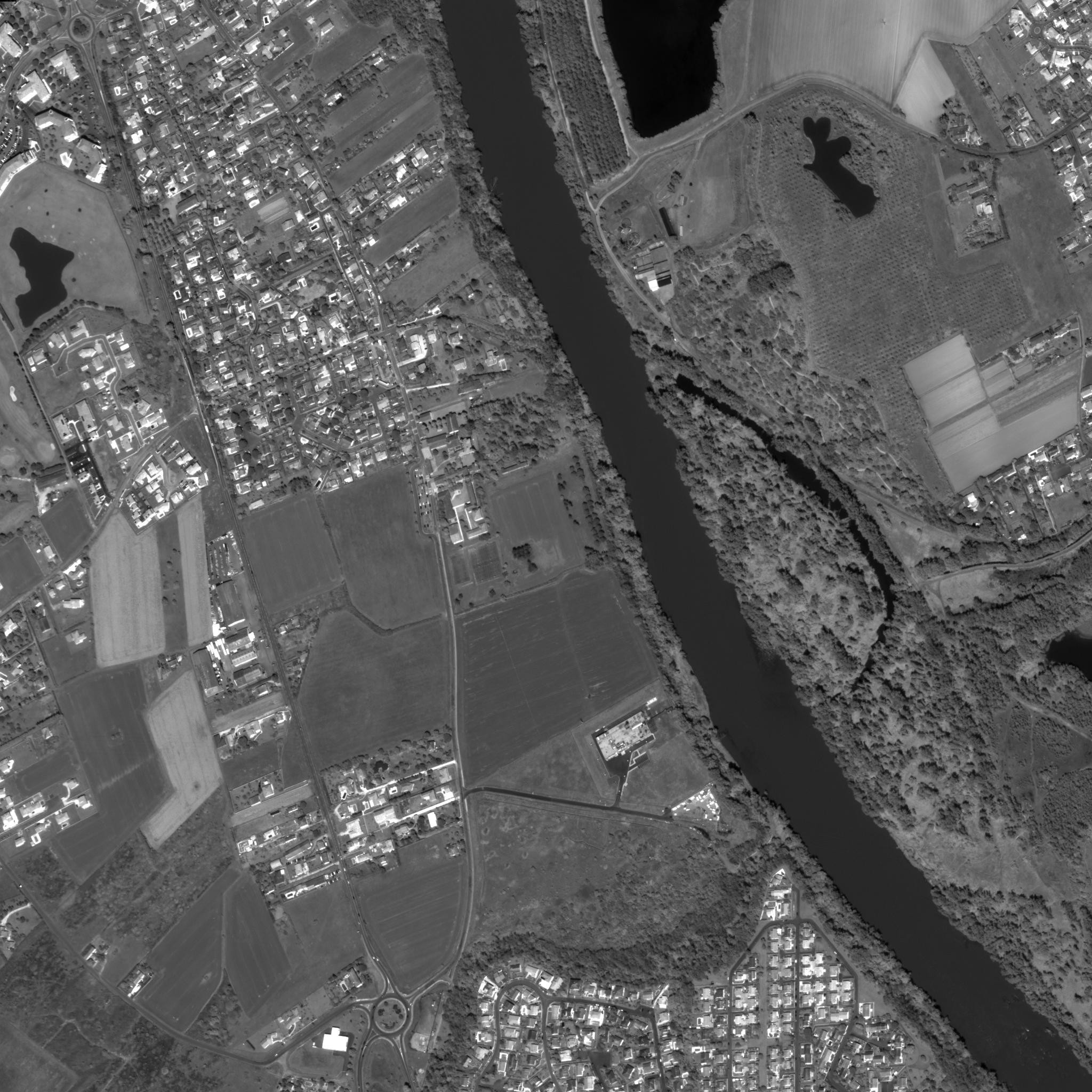} &
\includegraphics[trim= 8.7cm 38.1cm 56.2cm 26.8cm, clip=true, width=0.3\columnwidth]{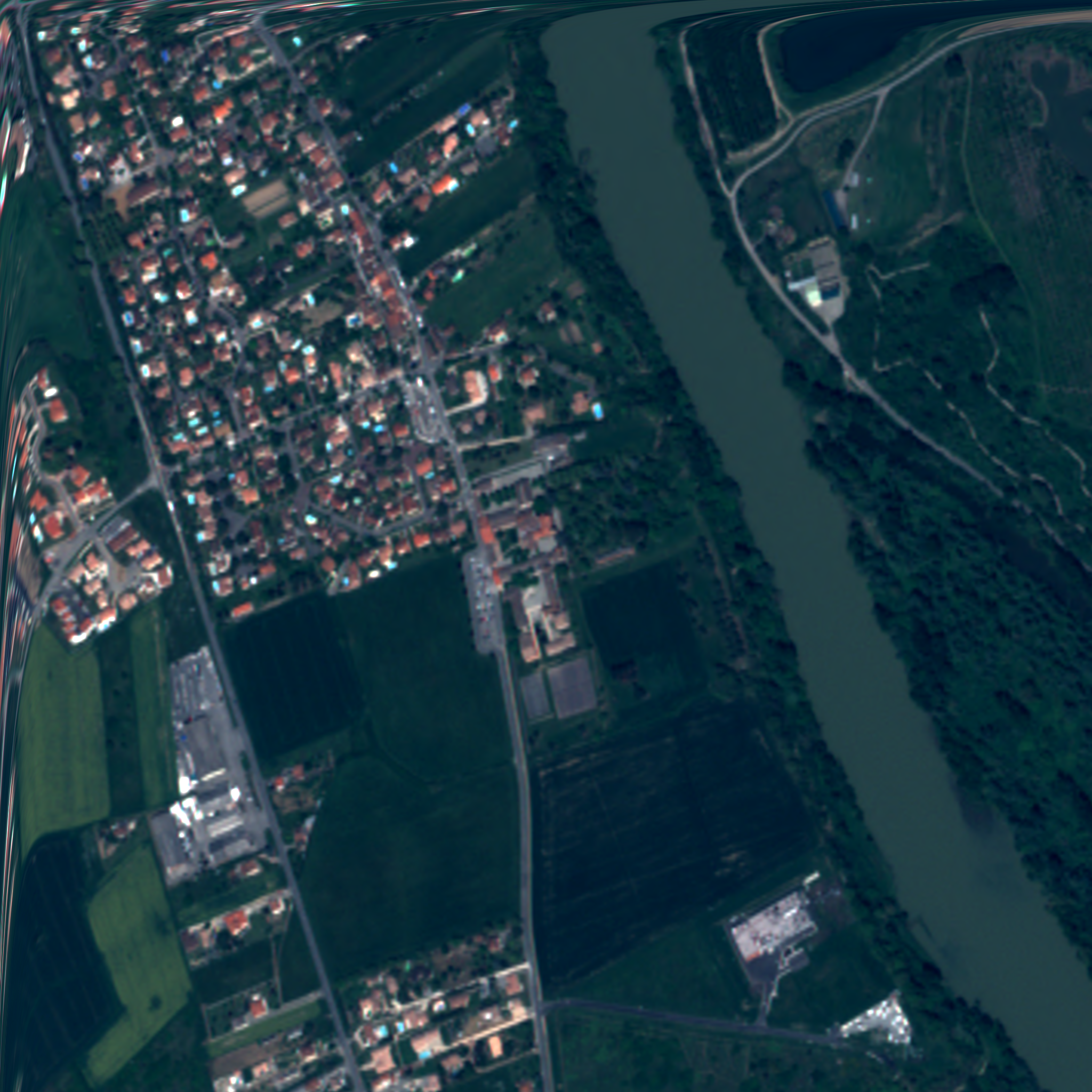}\\
Panchromatic & Multispectral
\end{tabular}
\caption{Data acquired by Pl{\'e}iades satellite on Toulouse (France), provided by CNES. The low-resolution multispectral image, which has been interpolated at the scale of the panchromatic, is seriously affected by aliasing.}
\label{fig_satellite_data}
\end{figure}

Many Earth observation satellites decouple the acquisition of a PAN image at high spatial resolution from that of a MS image at lower resolution (see Figure \ref{fig_satellite_data}). In remote sensing, having both high spatial and high spectral resolution is necessary for a wide range of applications, e.g., digital-surface model extraction and change detection. Pansharpening methods aim thus at producing a high-resolution MS image.
   
Satellite image fusion has been an intensive field of research \cite{LoncanFabre2015,VivoneAlparoneChanussot2015}. Component substitution (CS) \cite{ChavezSides1991, AiazziBarontiSelva2007, GarzelliNenciniCapobianco2008, ZhangMishra2014}, multi-resolution analysis (MRA) \cite{Liu2000, AiazziAlparoneBaronti2006, VivoneRestaino2014} and variational approaches (VAR) \cite{BallesterCaselles2006, DuranBuadesCollSbertSIIMS2014,DuranBuadesCollSbertBlanchetISPRS2017} have been mainly explored. CS methods are usually characterized by a high fidelity in rendering the spatial details but often suffer from significant spectral distortion, while MRA techniques are more successful in spectral preservation but experience spatial distortions \cite{VivoneAlparoneChanussot2015}.

Two of the most relevant challenges related to satellite acquisition systems are the misregistration between PAN and MS data and the strong aliasing in the MS bands. The first phenomenon is caused by the differences between the spatial and/or temporal coordinate systems of PAN and MS sensors. The latter is related to the optical characteristics of the MS sensor and usually produces spectral distortions and jagged edges as shown in Figure \ref{fig_satellite_data}. Under general and likely assumptions, Baronti {\it et al.}~\cite{BarontiAiazzi2011} proved that CS methods are less sensitive than MRA to both phenomena. Despite this, the experimental analysis conducted in \cite{DuranBuadesCollSbertBlanchetISPRS2017, BarontiAiazzi2011} revealed that the fused products still suffer from spatial and spectral inaccuracies and, thus, may need to be restored.

Several pansharpening approaches include restoration in their workflow. Massip {\it et al.}~\cite{MassipBlanc2012} proposed to take into account the difference in the modulation transfer function (MTF) between PAN and MS. It is conceived as an add-on to existing fusion methods. Vivone {\it et al.}~\cite{VivoneSimoes2015}  developed an algorithm for estimating the relation between PAN and MS images without the need of designing a filter to match the real MTF. The authors model the estimation as a blind image deblurring variational problem. Palsson {\it et al.}~\cite{PalssonSveinsson2016} replaced the interpolated MS image, which is the starting point of CS and MRA techniques, by its deblurred version. In these cases, the restoration is carried out before fusion. On the contrary, Lee and Lee \cite{LeeLee2010} introduced a post-processing stage to correct color distortions and noise near the edges. 

In this paper, we introduce a restoration method to improve the quality of pansharpened images. PCA transform is first applied to separate geometry and chromaticity. We propose a nonlocal variational model for filtering the {\it chromatic} components conditionally to the geometry of PAN. The geometric component is then replaced by the locally histogram-matched PAN for spatial enhancement. The full method does not require knowledge on the pansharpening approach and does not make use of the MS (aliased) data.  The proposed restoration scheme might not only remove possible errors of the pansharpened images, but also noise from the low-resolution MS bands.

The rest of the paper is organized as follows. Section \ref{sec_method} details the proposed post-processing chain, the performance of which is evaluated in Section \ref{sec_results} on fused products derived from different state-of-the-art pansharpening methods. Finally, conclusions are drawn in Section \ref{sec_conclusions}.

\section{Proposed Method} \label{sec_method}

We consider any image be given in a regular Cartesian grid and then rasterized by rows in a vector of $\R^N$, where $N$ is the number of pixels. We denote by $i\in\{1,\ldots, N\}$ the linearized index related to pixel coordinates $x_i=(x_i^1,x_i^2)$. Let $\P\in\R^N$ and $\f\in\R^{N\times M}$ respectively be the PAN and the fused image with $M$ spectral bands. We write $f_S, f_{C_1}, \ldots, f_{C_{M-1}}$ for the result of applying the PCA transform to $\f$, where $f_S$ is the geometric component and $f_{C_1}, \ldots, f_{C_{M-1}}$ the chromatic ones. We propose a nonlocal variational model to filter $f_{C_1}, \ldots, f_{C_{M-1}}$ conditionally to $\P$.  The resulting filtered components are denoted by $\tilde{f}_{C_1}, \ldots, \tilde{f}_{C_{M-1}}$. We then replace $f_S$ by the locally histogram-matched PAN, denoted by $\tilde{f}_S$. Finally, the transform is applied back to get the restored image $\tilde{\f}\in\R^{N\times M}$. 

\subsection{PCA Transform}\label{sec_transform}

We first apply the PCA transform \cite{Jolliffe2002} to $\f$. PCA is a simple, non-parametric method that uses an orthogonal transformation to convert a possibly confusing dataset to a new coordinate system in which the resulting vectors are linearly uncorrelated and follow modes of greatest variance in the data. The first principal component (PC), having the largest variance, is supposed to contain the structural information. 

\begin{figure}[!t]
\footnotesize
\centering
\begin{tabular}{c@{\hskip 0.02in}c@{\hskip 0.02in}c@{\hskip 0.02in}c@{\hskip 0.02in}c}
  \rotatebox{90}{\parbox[t]{0.75in}{\hspace*{\fill}1st PC\hspace*{\fill}}} &
  \includegraphics[trim= 21cm 31cm 15cm 5cm, clip=true, width=0.2\textwidth]{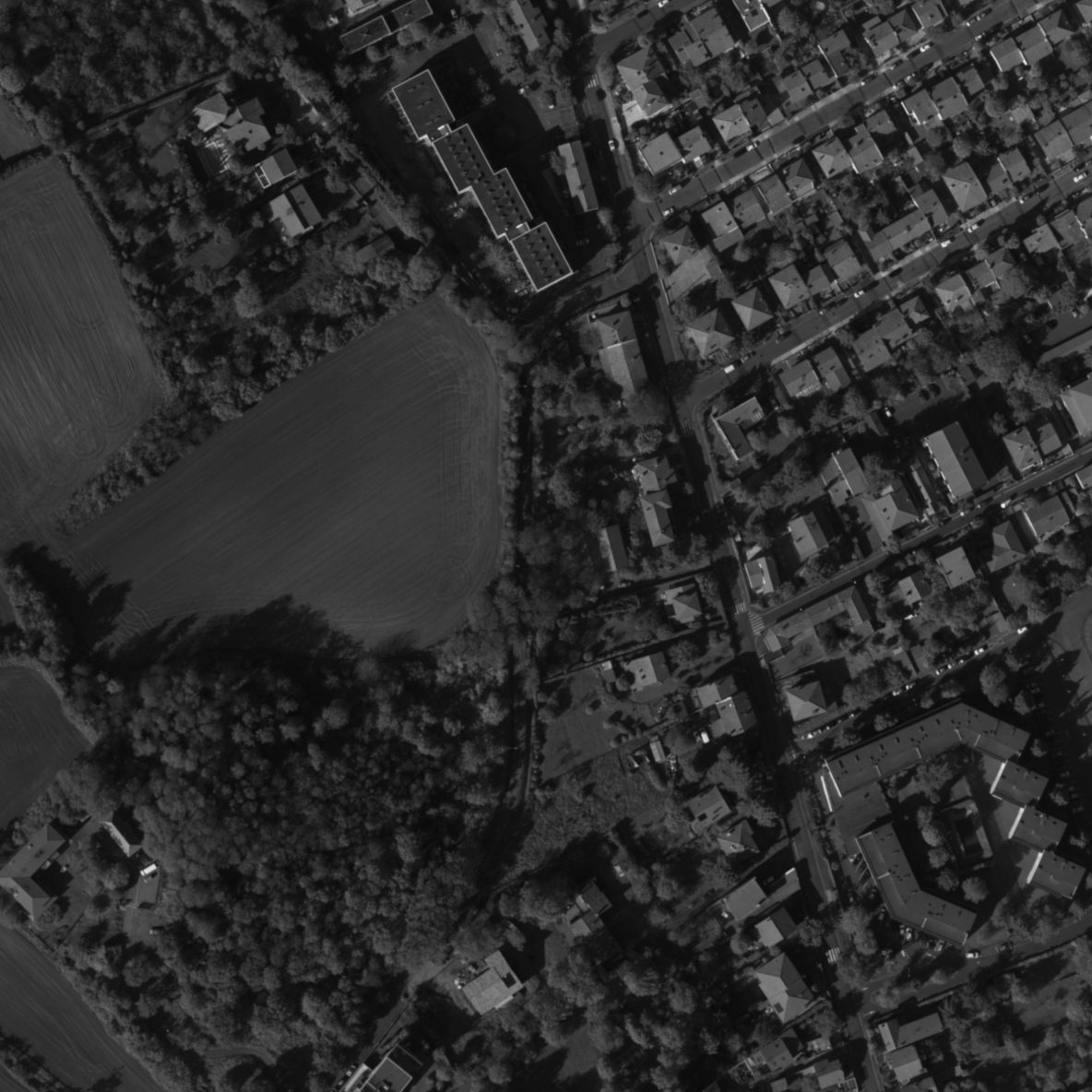} &
  \includegraphics[trim= 21cm 31cm 15cm 5cm, clip=true, width=0.2\textwidth]{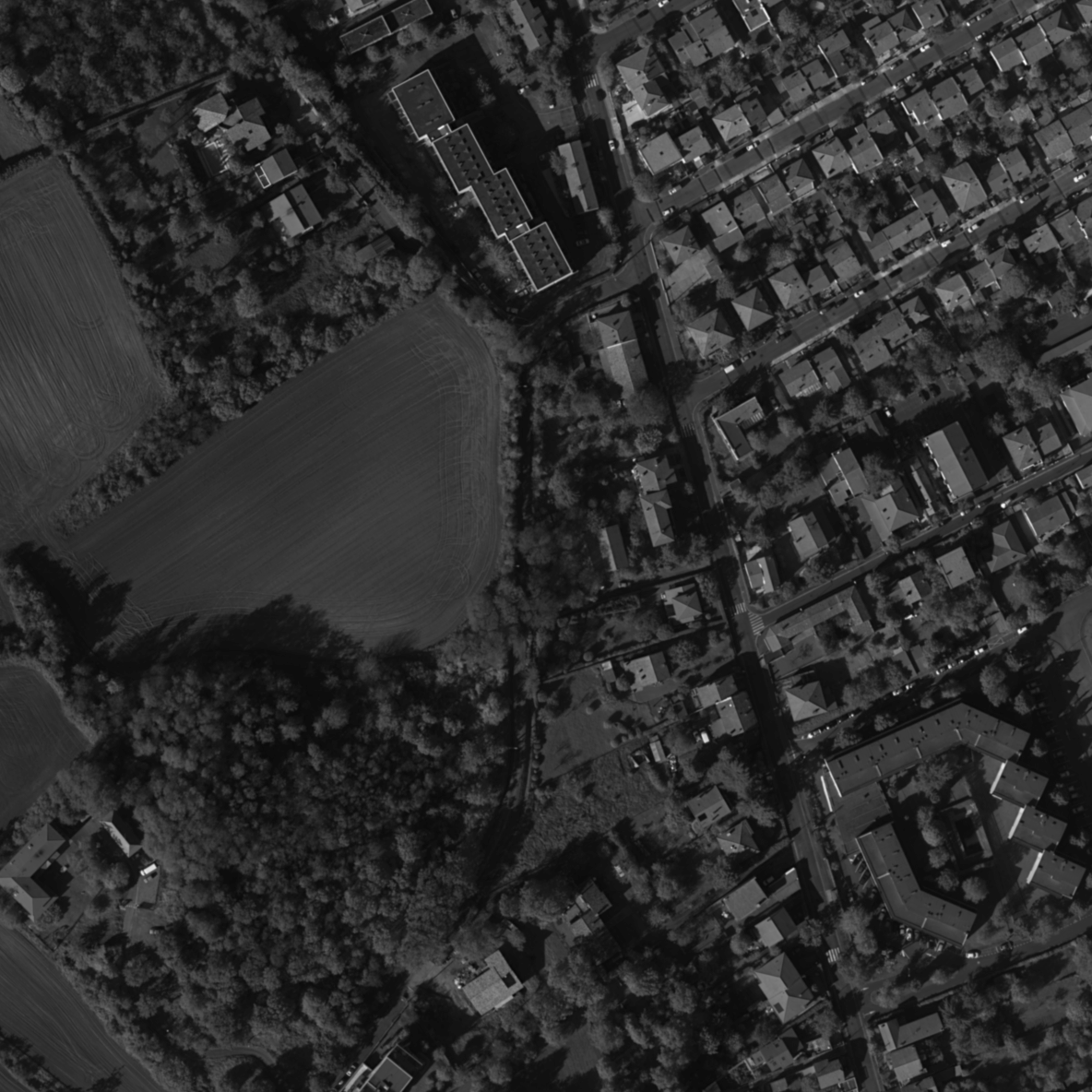} &
  \includegraphics[trim= 21cm 31cm 15cm 5cm, clip=true, width=0.2\textwidth]{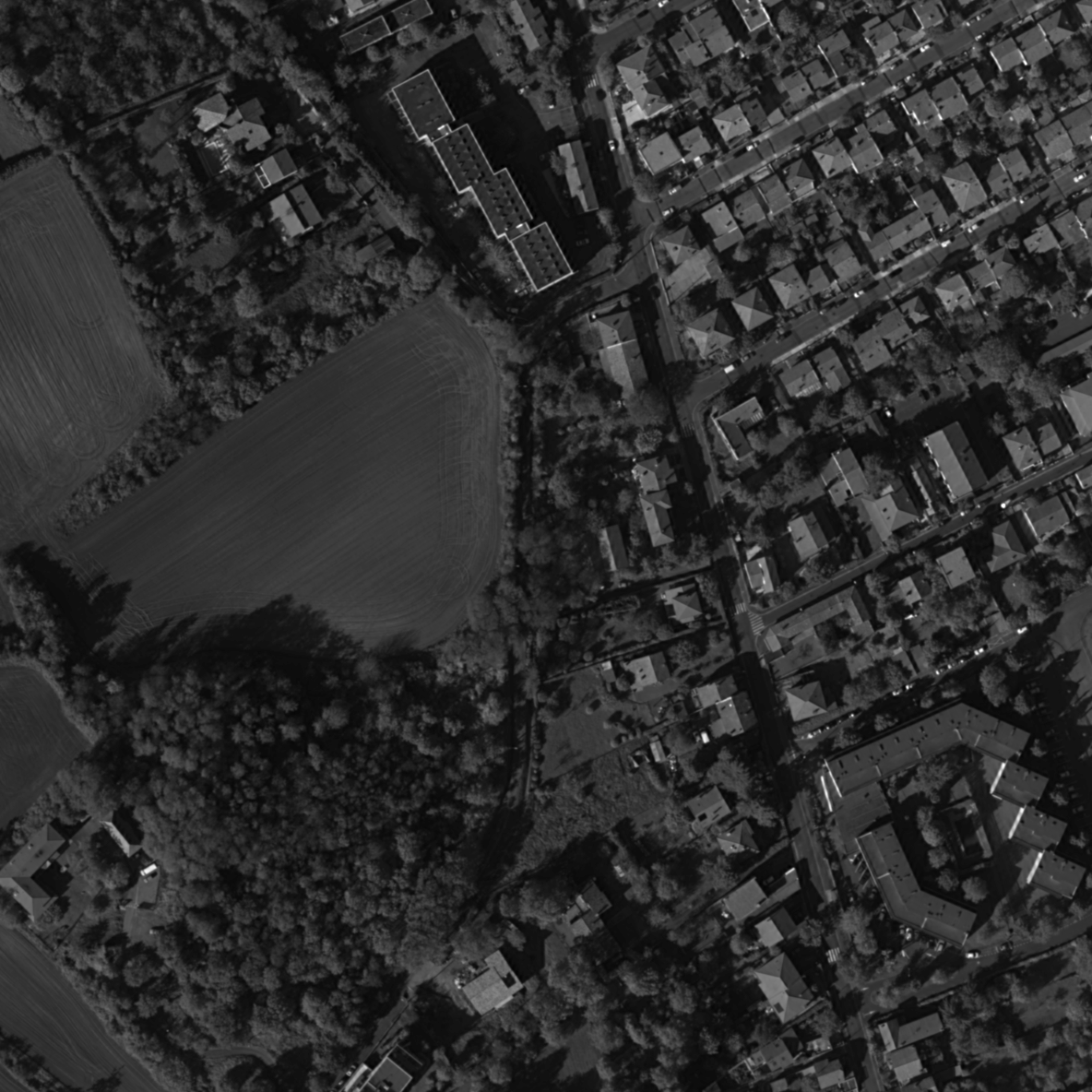} &
  \includegraphics[trim= 21cm 31cm 15cm 5cm, clip=true, width=0.2\textwidth]{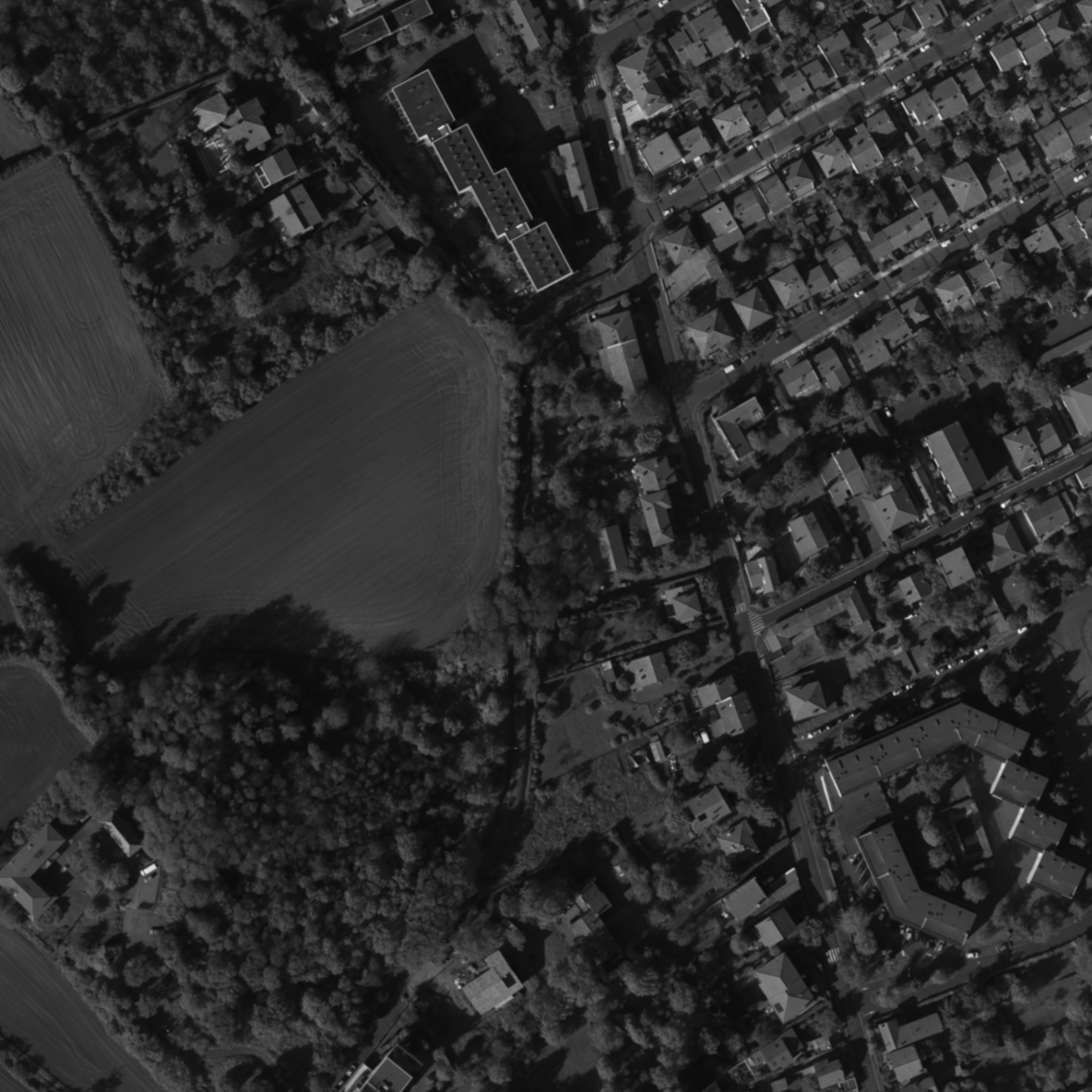} \\
  \rotatebox{90}{\parbox[t]{0.75in}{\hspace*{\fill}2nd PC\hspace*{\fill}}} &
  \includegraphics[trim= 21cm 31cm 15cm 5cm, clip=true, width=0.2\textwidth]{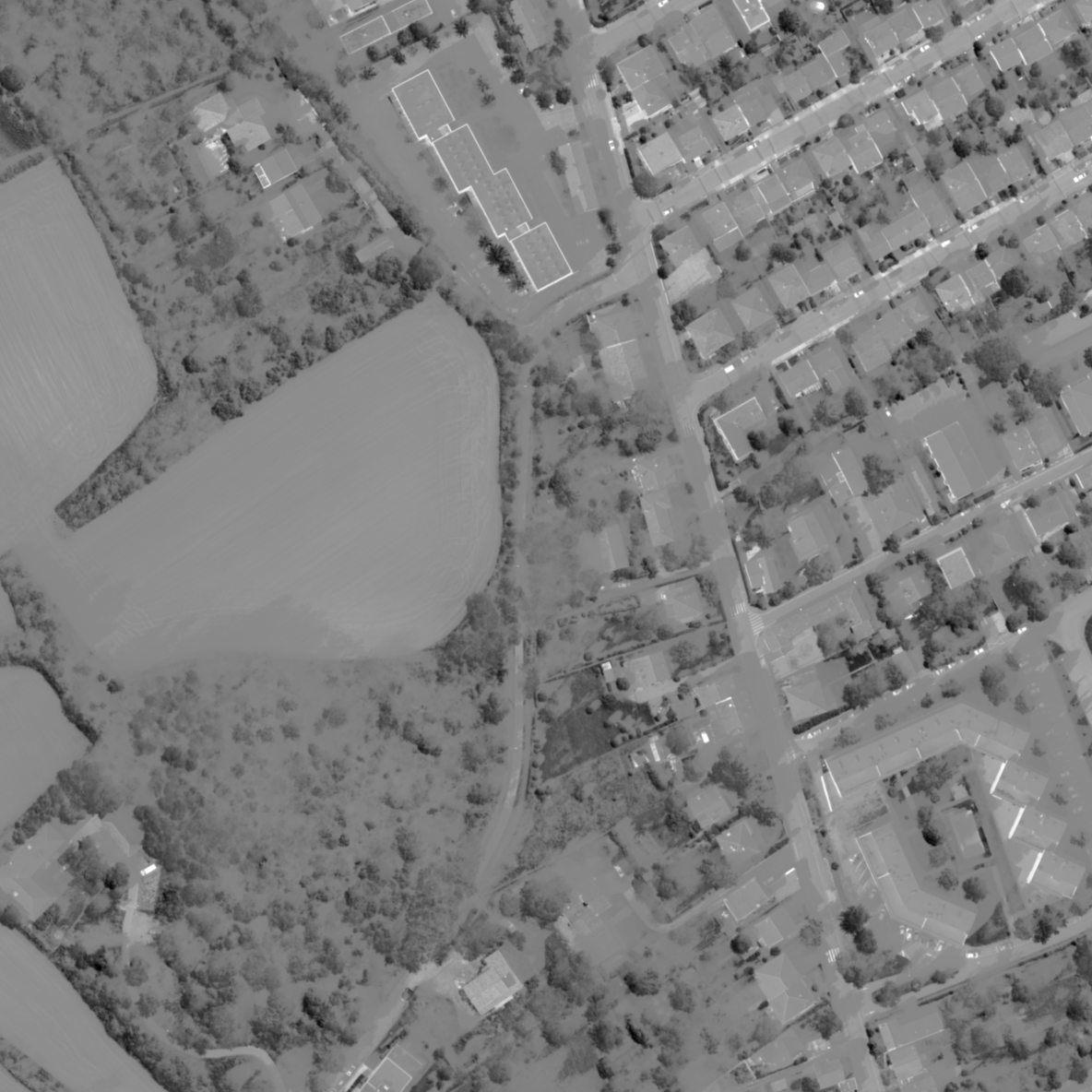} &
  \includegraphics[trim= 21cm 31cm 15cm 5cm, clip=true, width=0.2\textwidth]{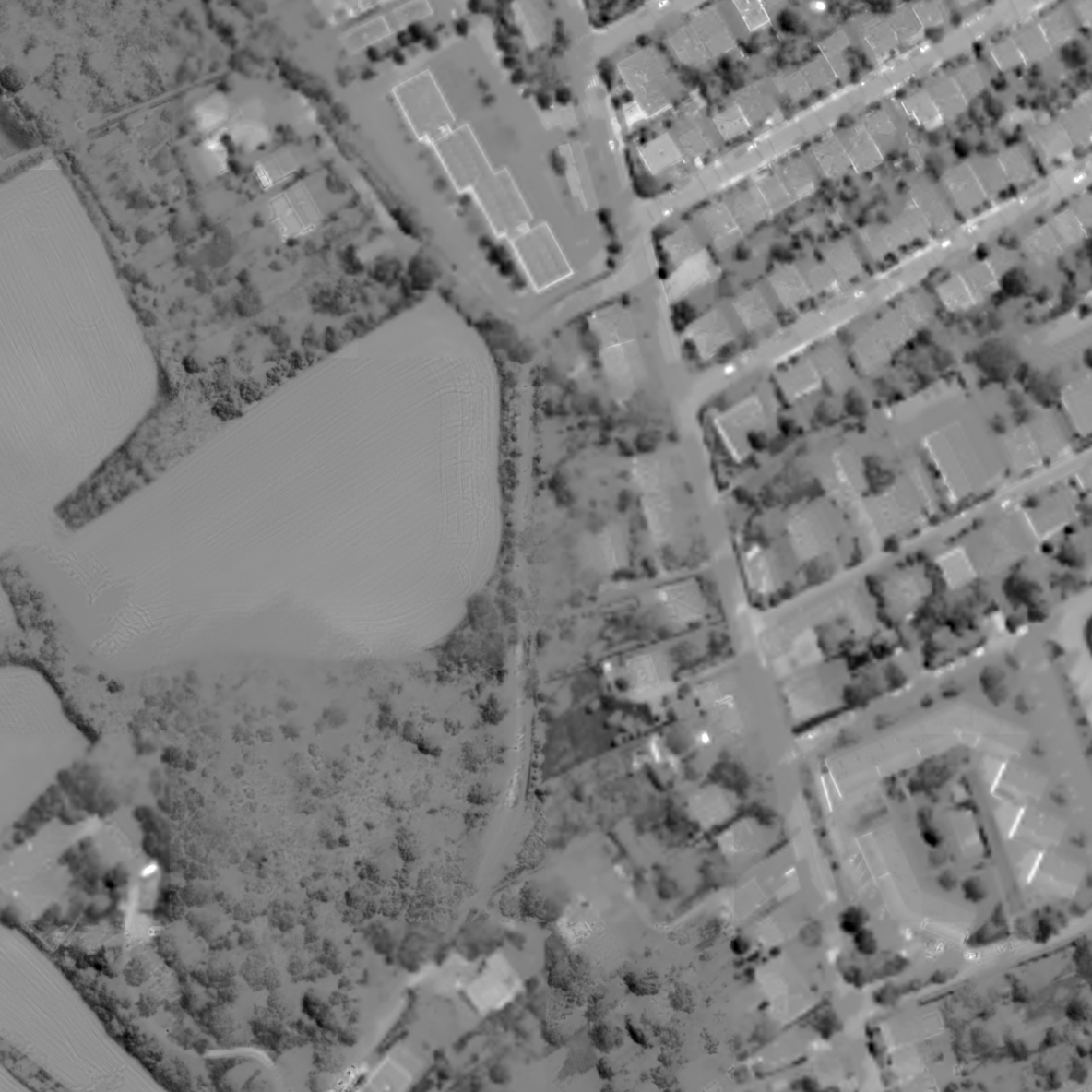} &
  \includegraphics[trim= 21cm 31cm 15cm 5cm, clip=true, width=0.2\textwidth]{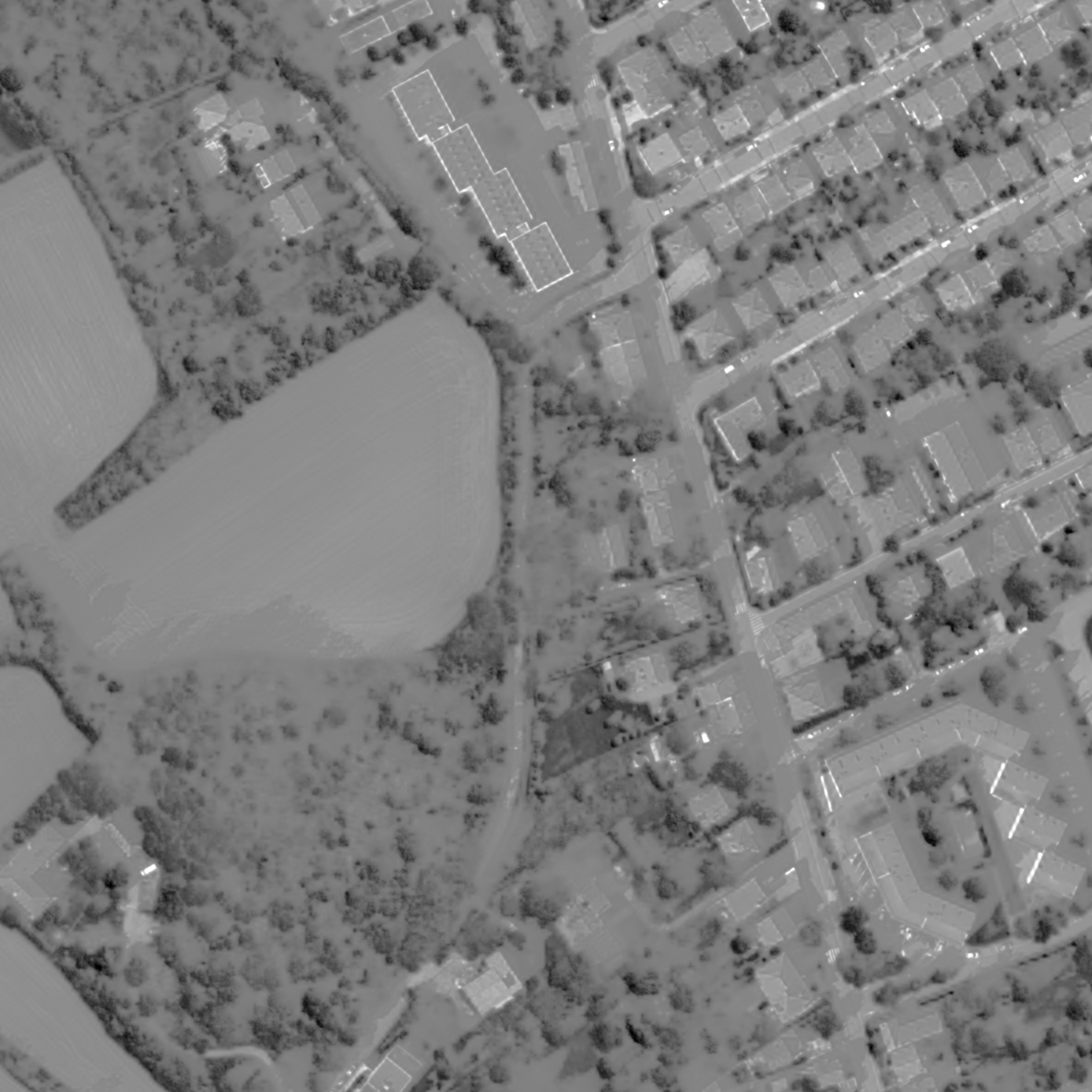} &
  \includegraphics[trim= 21cm 31cm 15cm 5cm, clip=true, width=0.2\textwidth]{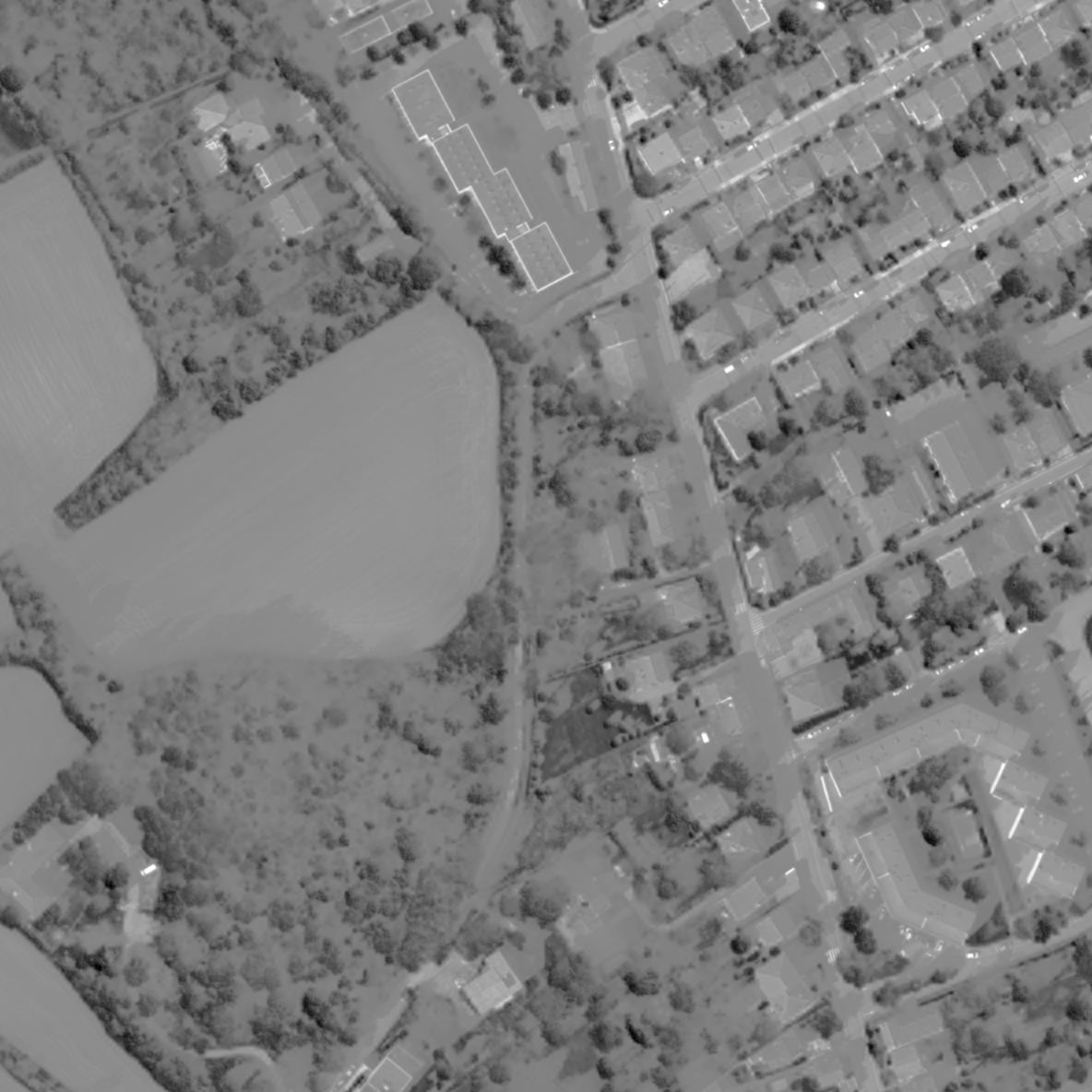} \\
  \rotatebox{90}{\parbox[t]{0.75in}{\hspace*{\fill}3rd PC\hspace*{\fill}}} &
  \includegraphics[trim= 21cm 31cm 15cm 5cm, clip=true, width=0.2\textwidth]{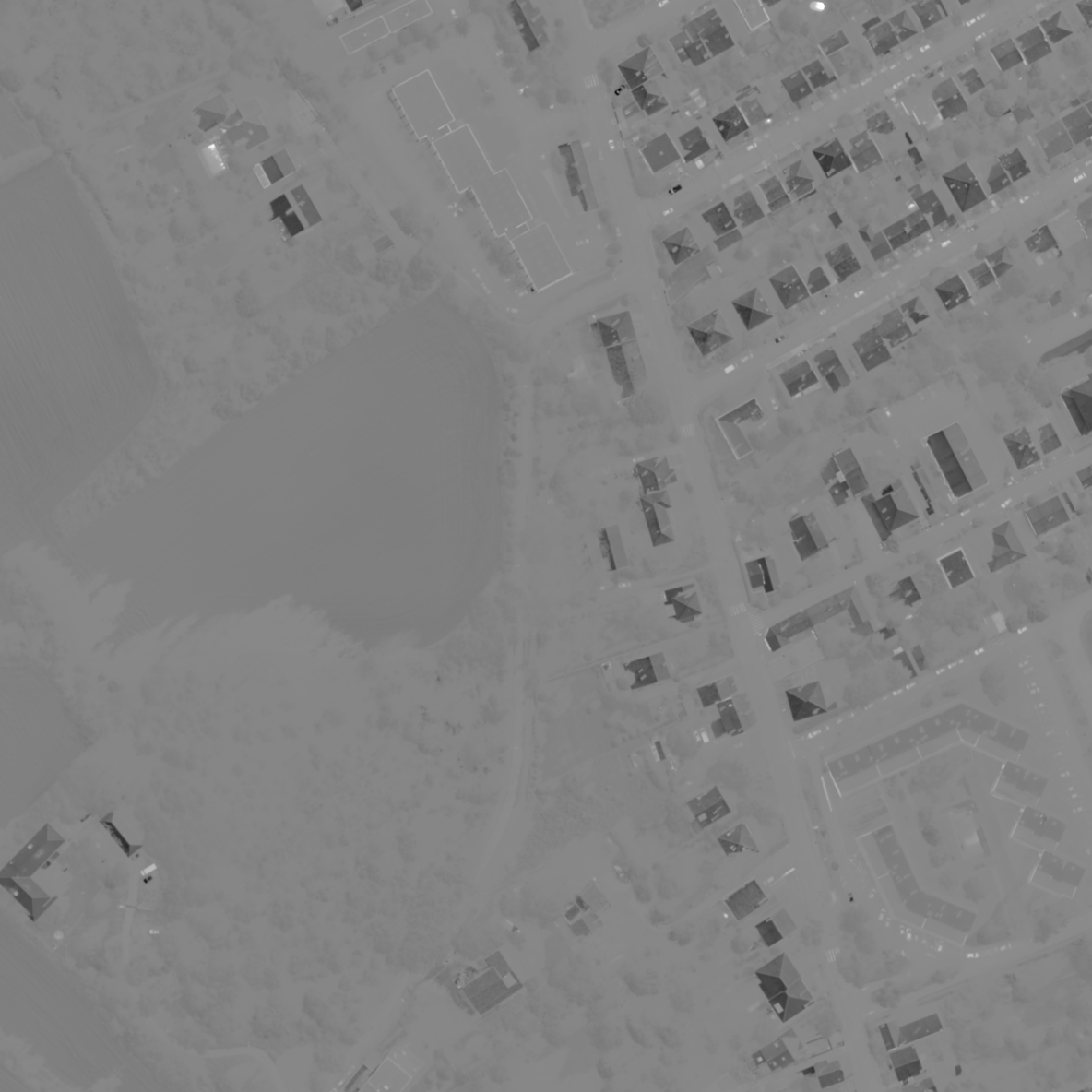} &
  \includegraphics[trim= 21cm 31cm 15cm 5cm, clip=true, width=0.2\textwidth]{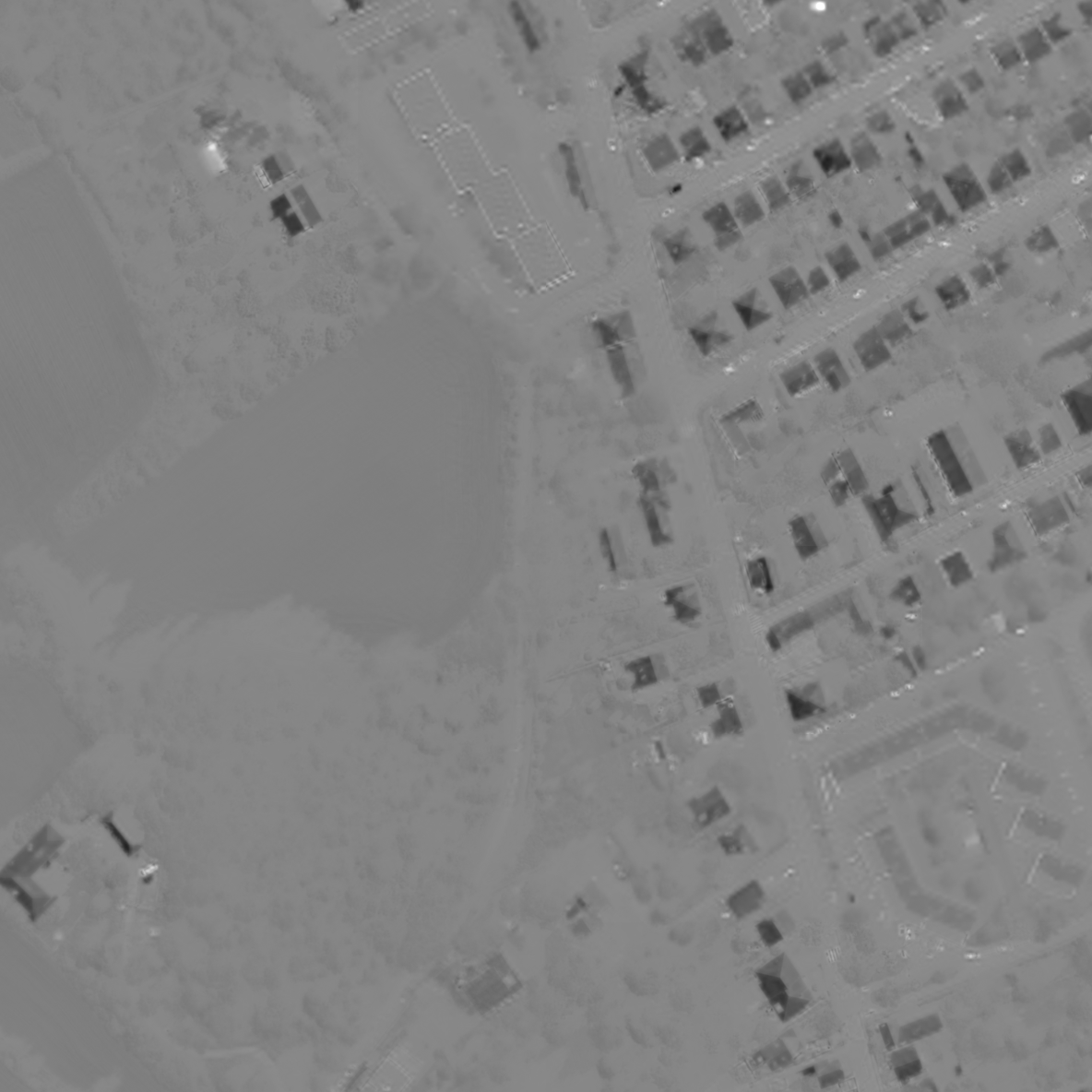} &
  \includegraphics[trim= 21cm 31cm 15cm 5cm, clip=true, width=0.2\textwidth]{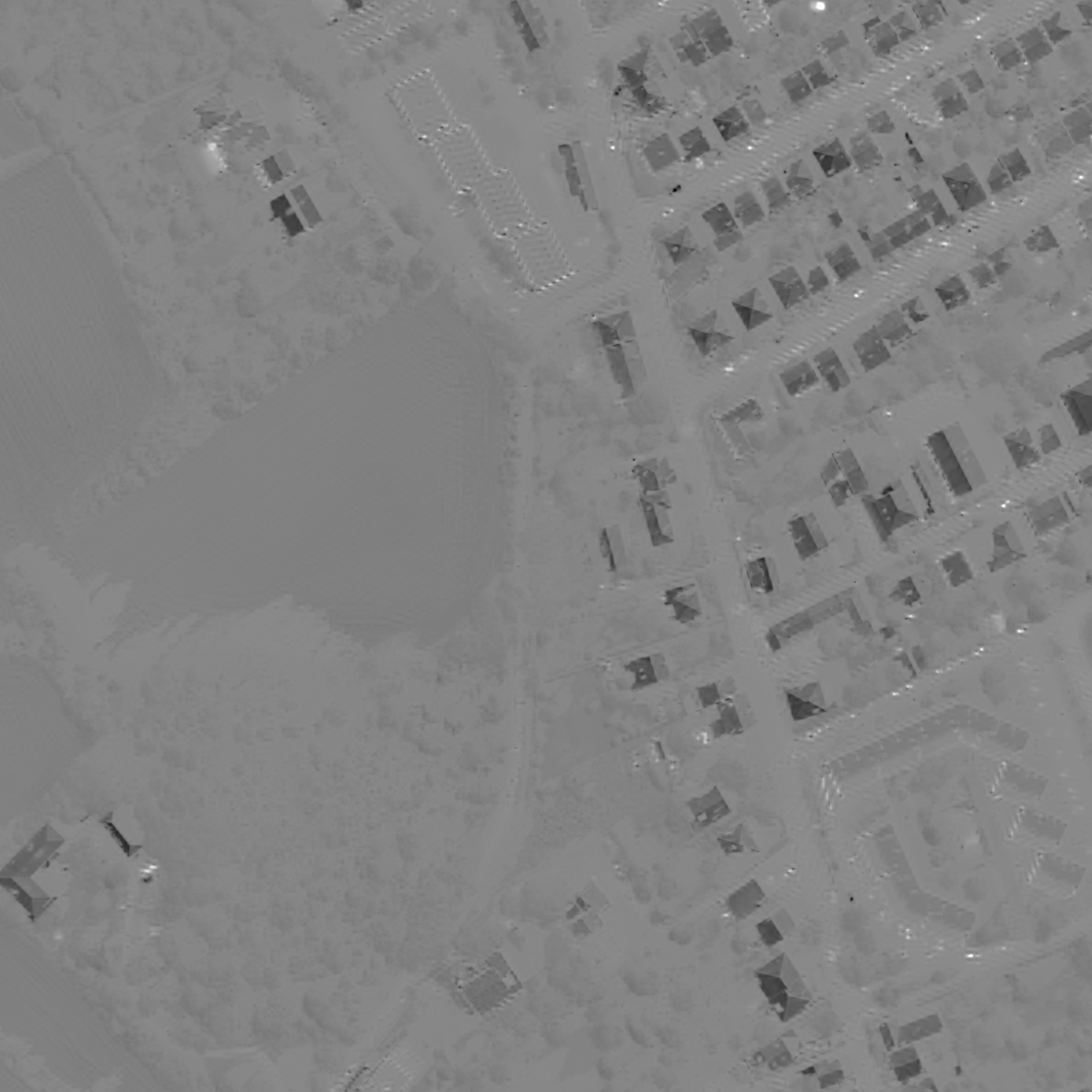} &
  \includegraphics[trim= 21cm 31cm 15cm 5cm, clip=true, width=0.2\textwidth]{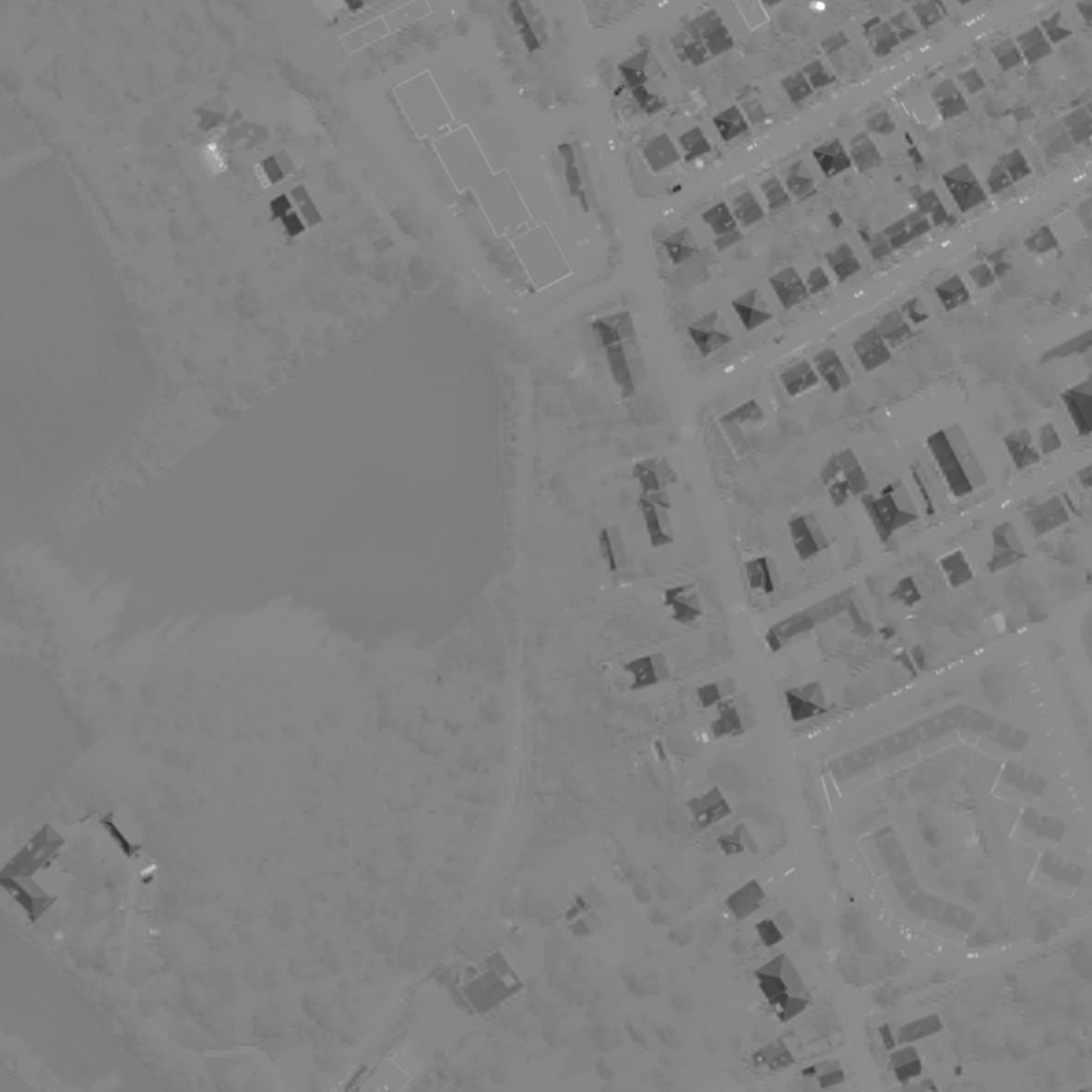} \\
  \rotatebox{90}{\parbox[t]{0.75in}{\hspace*{\fill}4rd PC\hspace*{\fill}}} &
  \includegraphics[trim= 21cm 31cm 15cm 5cm, clip=true, width=0.2\textwidth]{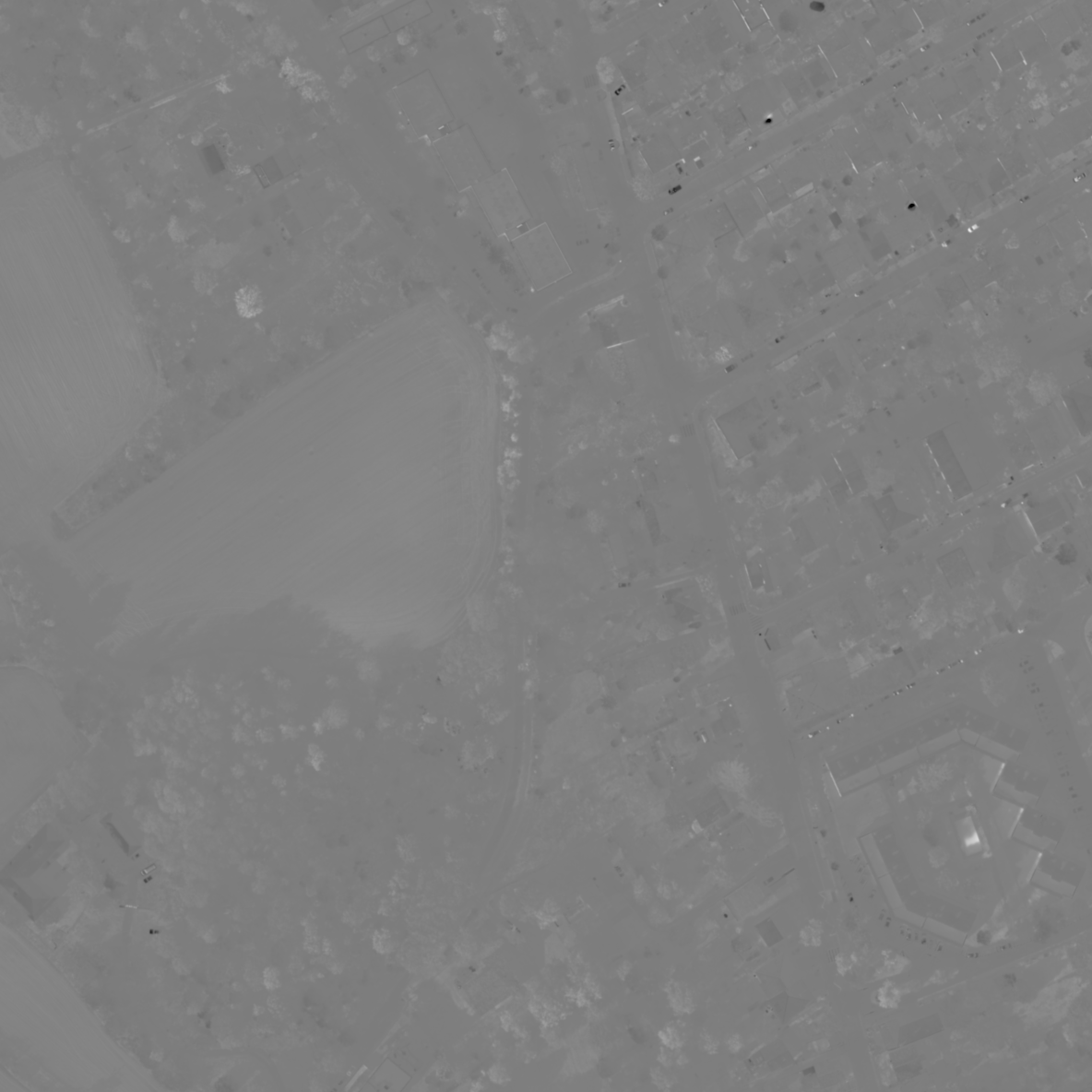} &
  \includegraphics[trim= 21cm 31cm 15cm 5cm, clip=true, width=0.2\textwidth]{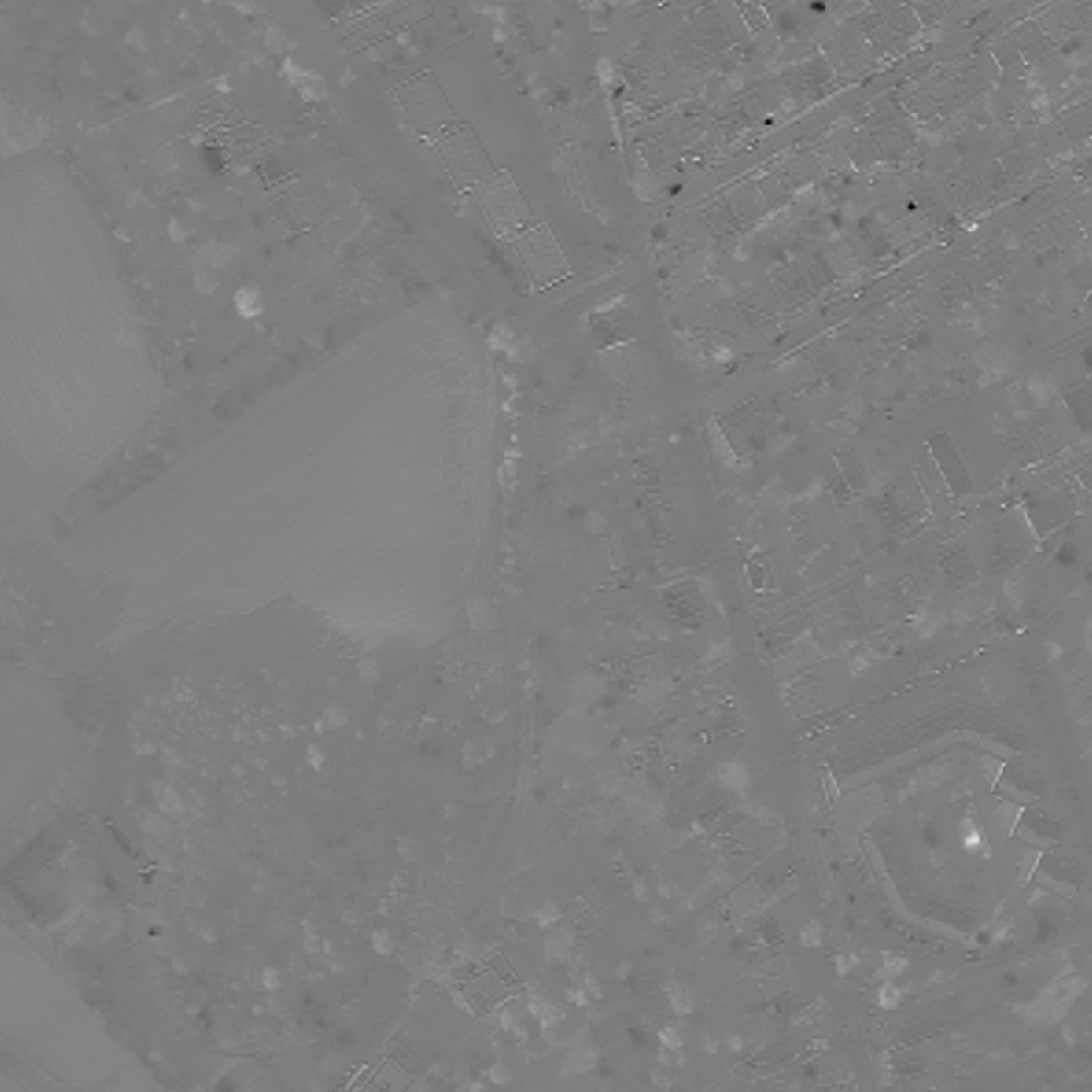} &
  \includegraphics[trim= 21cm 31cm 15cm 5cm, clip=true, width=0.2\textwidth]{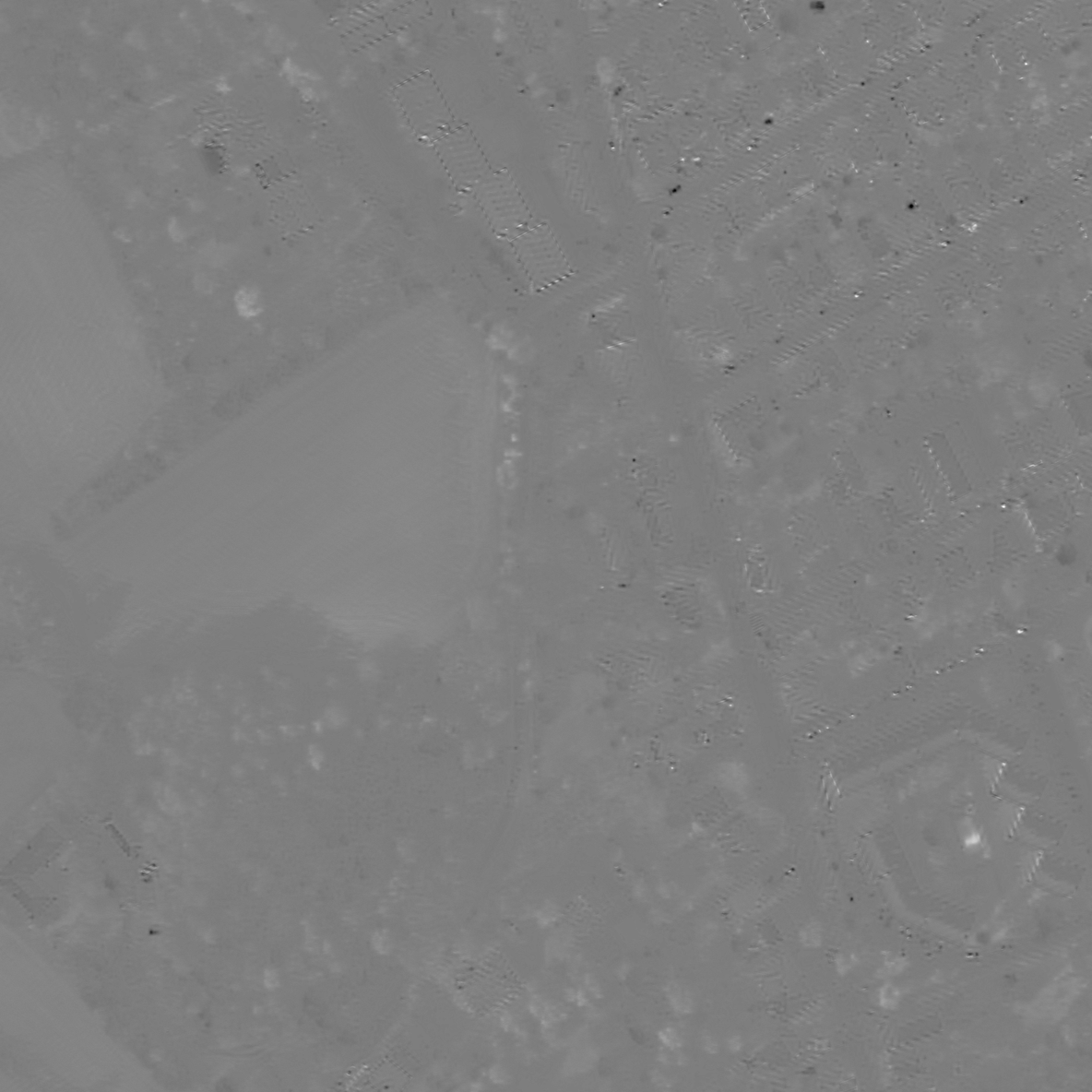} &
  \includegraphics[trim= 21cm 31cm 15cm 5cm, clip=true, width=0.2\textwidth]{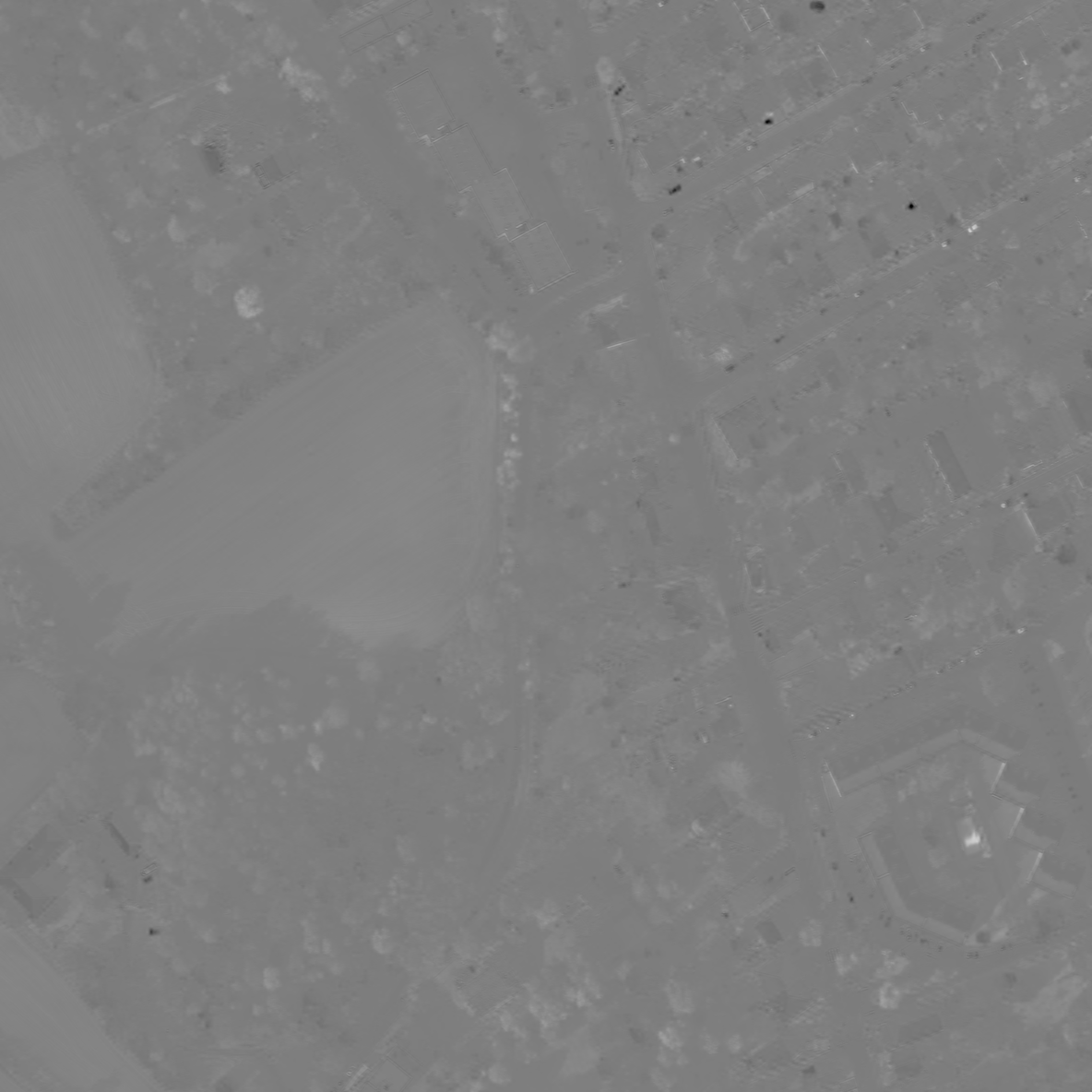} \\
  &Reference & BDSD \cite{GarzelliNenciniCapobianco2008} & GLP \cite{AiazziAlparoneBaronti2006} & NLVD \cite{DuranBuadesCollSbertBlanchetISPRS2017}
\end{tabular}
\caption{Fused images after PCA and linear rescaling of the displayed intensities. For visualization purposes, a gamma correction of factor 0.75 has been applied to the images containing 1st PCs. Although some blurring emerge in the 1st PCs, most of the distortions are concentrated in the other PCs.}
\label{fig_artifacts_pca}
\end{figure}

Figure \ref{fig_artifacts_pca} displays the PCs obtained after applying the PCA transform to the pansharpened images by band dependent spatial detail (BDSD) \cite{GarzelliNenciniCapobianco2008}, generalized Laplacian pyramid with MTF matched filter and high-pass modulation (GLP) \cite{AiazziAlparoneBaronti2006} and band-decoupled nonlocal variational model (NLVD) \cite{DuranBuadesCollSbertBlanchetISPRS2017}. The PAN and MS data (blue, green, red and near infra-red) were generated following the procedure in Section \ref{sec_aerial}. We notice that  blur emerges in the structural components (1st PCs). However, most of the distortions such as jagged edges and drooling effects are concentrated in the other PCs. These results are consistent with the theoretical analysis in \cite{BarontiAiazzi2011} since the MRA-based technique GLP is very sensitive to aliasing, which is even apparent in the 1st PC. As shown in \cite{DuranBuadesCollSbertBlanchetISPRS2017}, NLVD compares favorably since artifacts in its image are only noticeable in the 3rd and 4rd PCs, which have small energy.

\subsection{Nonlocal Spectral Filtering}

We propose a nonlocal convex variational model for filtering the chromatic components. The cost function to be minimized includes a data-fidelity term preserving the original content up to a certain extent and a nonlocal regularization term conditioned to the geometry of PAN. The nonlocal prior \cite{BuadesCollMorel2005} relates pixels with similar geometry and texture characteristics. 

For $m\in\{1,\ldots, M\hspace{-0.06cm}-\hspace{-0.06cm}1\}$, the energy minimization writes as
\begin{equation}\label{eq_energy_chrominance}
\tilde{f}_{C_m} = \arg\min_{u\in\R^N} \lambda \left\|\nabla_{\omega} u\right\|_1 + \tfrac{1}{2}\left\|u - f_{C_m}\right\|_2^2,
\end{equation}
where $\lambda>0$ is a trade-off parameter. The nonlocal gradient $\nabla_{\omega} u\in\R^{N\times N}$, with $\nabla_{\omega} u_i=\left((\nabla_{\omega}u)_{i,1},\ldots,(\nabla_{\omega} u)_{i,N}\right)$ at each pixel, is given in terms of the weighted differences 
\begin{equation}
\left(\nabla_{\omega} u\right)_{i,j} = \sqrt{\omega_{i,j}} \left(u_j-u_i\right),
\end{equation}
where the weights $\omega_{i,j}$ measure the {\it similarity} between pixels. The gradient is penalized in \eqref{eq_energy_chrominance} as $\|\nabla_{\omega} u\|_1=\sum_i \left\|\nabla_{\omega} u_i\right\|$, where $\|\cdot\|$ denotes the Euclidean norm. The minimization applies to each band independently, thus each one is filtered with no contribution of the others. For simplicity, we assume the same $\lambda$ for all components. Better results might be obtained by looking for band-dependent trade-off parameters.

The weights are computed on the PAN image that describes accurately the geometry of the scene. They take into account both the spatial closeness and the patch similarity in $\P$. For computational purposes, the nonlocal gradient is limited to interact only between pixels at a certain distance, i.e.:
\begin{equation}\label{eq_nonlocal_weights_def}
\omega_{i,j} = \dfrac{1}{\Gamma_i} \exp\left( -\dfrac{ \|x_i-x_j\|^2}{h^2_{\text{spt}}} - \dfrac{ \left\|\P(p_i) - \P(p_j)\right\|^2}{h^2_{\text{sim}}}\right) 
\end{equation}
if $\|x_i-x_j\|_{\infty}\leq \nu_r$ and zero otherwise. In this setting, $\nu_r\in\Z^+$ determines the size of the window where to search for similar pixels and $\P(p_i)$ denotes a patch in PAN centered at pixel coordinates $x_i$. The normalization factor $\Gamma_i$ is given by
\begin{equation}\label{eq_nonlocal_weights_norm}
\Gamma_i = \hspace{-0.04cm}\sum_{x_j\in\mathcal{N}_i}\exp\left( -\dfrac{\|x_i-x_j\|^2}{h^2_{\text{spt}}} - \dfrac{ \left\|\P(p_i) - \P(p_j)\right\|^2}{h^2_{\text{sim}}}\right),
\end{equation}
where $\mathcal{N}_i=\{x_j \mid \|x_i-x_j\|_{\infty}\leq \nu_r\}$. The filtering parameters $h_{\text{spt}}$ and $h_{\text{sim}}$ measure how fast the weights decay with increasing spatial distance or dissimilarity between patches, respectively. In practice, the weight distribution is in general sparse since a few nonzero values are considered, thus the nonlocal gradient reduces to $\nabla_{\omega}u\in\R^{N\times (2\nu_r+1)^2}$.

The problem \eqref{eq_energy_chrominance} is convex but non smooth. To find a fast, global optimal solution we use the first-order primal-dual algorithm \cite{ChambollePock2011}, which was proposed for solving saddle-point problems. Since the Legendre-Fenchel transform of a norm is the indicator function of the unit dual norm ball, the primal problem \eqref{eq_energy_chrominance} can be rewritten in a saddle-point formulation as
\begin{equation}\label{eq_chrominance_pd}
\min_{u} \max_{q} \, \langle \nabla_{\omega} u, q\rangle - \delta_{\mathcal{Q}}(q) + \tfrac{1}{2}\left\|u-f_{C_m}\right\|_2^2,
\end{equation}
where $q\in\R^{N\times(2\nu_r+1)^2}$ is the dual variable related to the nonlocal regularization and $\delta_{\mathcal{Q}}$  is the indicator function of the convex set $\mathcal{Q}=\lbrace q : \|q\|_{\infty}\leq \lambda\rbrace$, with $\|q\|_{\infty}=\max_i \|q_i\|$.

The efficiency of the algorithm depends on the proximity operators having closed-form representations or being solved with high precision. In our case, the proximity operators of $G(u):=\frac{1}{2}\|u-f_{C_m}\|_2^2$ and $F^{*}(q):=\delta_{\mathcal{Q}}(q)$ are, respectively, $\prox_{\tau G}(u) = \frac{u+\tau f_{C_m}}{1+\tau}$ and $(\prox_{\sigma F^{*}} (q))_{i,j} = \tfrac{\lambda q_{i,j}}{\max\left\{\lambda,\|q_i\|\right\}}$.  The solution of \eqref{eq_chrominance_pd} is thus computed as in \cite[Algorithm 1]{ChambollePock2011}.

\begin{figure}[!t]
\centering \renewcommand{\arraystretch}{0.5}
\begin{tabular}{c@{\hskip 0.03in}c@{\hskip 0.03in}c@{\hskip 0.03in}c@{\hskip 0.03in}c@{\hskip 0.03in}c} 
  \includegraphics[trim= 20cm 6cm 9.3cm 22.9cm, clip=true, width=0.23\columnwidth]{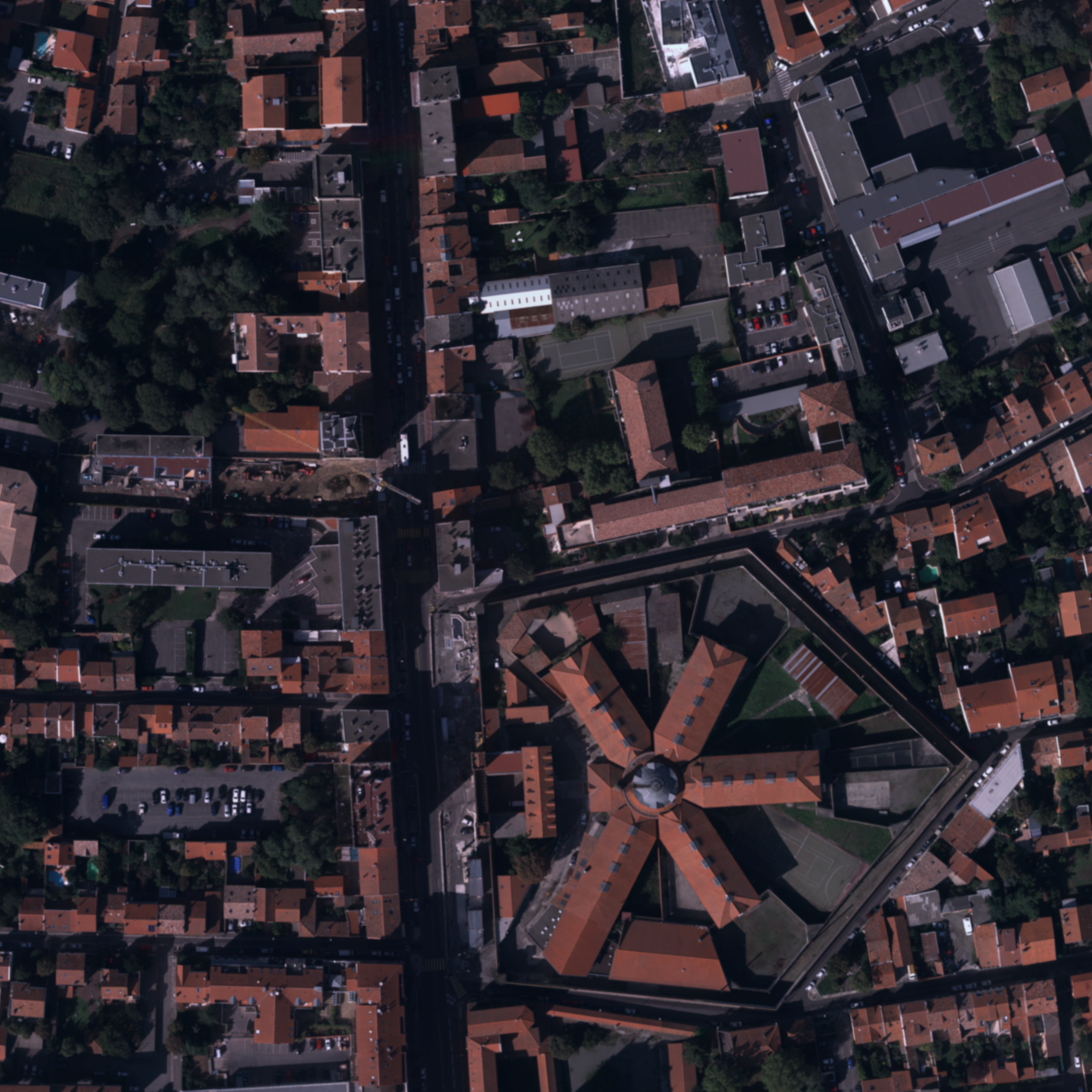} &
  \includegraphics[trim= 0cm 28.1cm 29cm 0.5cm, clip=true, width=0.23\columnwidth]{pelican1_gamma.png} &
  \includegraphics[trim= 23.3cm 20.7cm 6.3cm 8.5cm, clip=true, width=0.23\columnwidth]{pelican1_gamma.png} \\ 
  \includegraphics[trim= 6.5cm 25.8cm 23cm 3.3cm, clip=true, width=0.23\columnwidth]{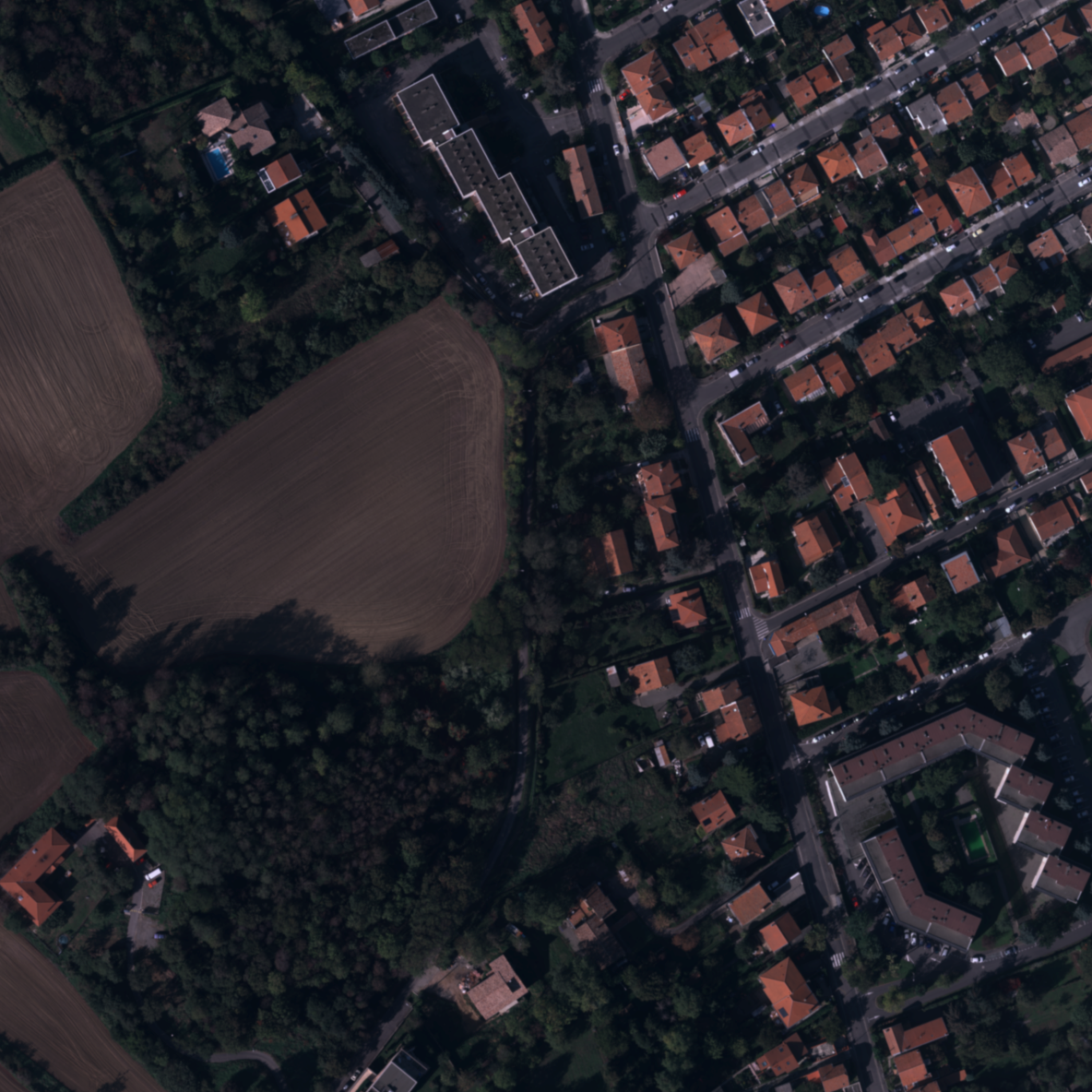} &
  \includegraphics[trim= 29.5cm 2.8cm 0cm 26.3cm, clip=true, width=0.23\columnwidth]{pelican2_gamma.png} &
  \includegraphics[trim= 21.4cm 27cm 8cm 2cm, clip=true, width=0.23\columnwidth]{pelican2_gamma.png} 
\end{tabular}
\caption{Ground-truth 4-band images of size $512\times 512$ and resolution 30 cm per pixel, courtesy of CNES. For visualization purposes, a gamma correction of factor 0.75 has been applied.}
\label{fig_pelican}
\end{figure}

\subsection{Structural Component Replacement}\label{sec_hist_pan}

The structural component often suffers from blurring effects. Inspired by CS pansharpening, we propose to replace $f_S$ with PAN in order to recover spatial resolution.  Histogram matching of $\P$ to $f_S$ is usually performed so that the mean and standard deviation of $f_S$ and the modified PAN are within the same range. Hence, the enhanced structural component would be
\begin{equation}\label{eq_hist_global}
\tilde{f}_S=\dfrac{\sigma_{f_S}}{\sigma_{\P}}\left(\P-\mu_{\P}\right) +\mu_{f_S},
\end{equation}
where $\mu$ and $\sigma$ denote global mean and standard deviation.

The above approach does not account for local illumination dissimilarities between $\P$ and $f_S$ since operates on the whole images.  In order to deal with this issue, we use local histogram matching. We pan across the images using sliding patches and perform locally the equalization in \eqref{eq_hist_global}.  Since patches overlap, all assigned values per pixel are aggregated by averaging.  

If the size of the patch is too small, such a local histogram matching will not only transfer the local intensity distribution of the geometric component  but its entire content. If we reduce the patch to a single pixel, equation  \eqref{eq_hist_global} amounts to keep the geometrical component instead of a matched PAN. The proposed structural replacement is less effective for CS methods, since they already apply a global matching and substitution by PAN.

\section{Experimental Results} \label{sec_results}

In this section, we evaluate the performance of the proposed method for restoring pansharpened images. We use fused products by the CS methods Gram-Schmidt adaptive (GSA) \cite{AiazziBarontiSelva2007} and BDSD \cite{GarzelliNenciniCapobianco2008}, the MRA techniques additive wavelet luminance proportional (AWLP) \cite{OtazuGonzalezFors2005} and GLP \cite{AiazziAlparoneBaronti2006}, and the VAR approaches Bayesian sparse pansharpening (BAY) \cite{LoncanFabre2015} and NLVD \cite{DuranBuadesCollSbertBlanchetISPRS2017}. The source codes of all methods except NLVD were obtained from the toolboxes in \cite{LoncanFabre2015, VivoneAlparoneChanussot2015}.

In our algorithm, we fixed  $\nu_r=7$, i.e.~$15\times 15$ research windows are used as a compromise between accuracy and computational efficiency. This research window should be increased for higher resolution images in order to account for the same surface area. We also set the size of the patches $\P(p_i)$ to $3\times 3$ and $h_{\text{spt}}=2.5$. We defined $15\times 15$ sliding patches and one pixel as sliding distance for the local histogram matching  described in Section \ref{sec_hist_pan}. The remaining parameters $h_{\text{sim}}$ and $\lambda$ were chosen by running the algorithm on the dataset from Figure \ref{fig_pelican} for several random values and picking up the optimal ones in terms of the RMSE.

The proposed restoration method was run on a laptop with one core Processor 2.5 GHz Intel Core i5 with 4 GB 1600 MHz DDR3 RAM. For the restoration of a 4-band image of size $512\times 512$, the running time required was 18.23 seconds.  Several parts of the algorithm could be easily parallelized. Patch comparison during the weight computation could be accelerated by using pre-selection strategies  \cite{MahmoudiSapiro2005}.

\begin{table}[t]
\centering
\centering
{\small
\setlength\tabcolsep{2.6pt}
\renewcommand{\arraystretch}{1.05}
\begin{tabular}{|c|c|c|c|c|c|c|c|c|}
\hline
\multirow{2}{*}{Methods} & \multicolumn{2}{c|}{RMSE} & \multicolumn{2}{c|}{ERGAS} & \multicolumn{2}{c|}{SAM} & \multicolumn{2}{c|}{Q$2^n$}  \\ \cline{2-9}
 & Fus & Rest & Fus & Rest & Fus & Rest & Fus & Rest \\ \hline\hline
Reference & \multicolumn{2}{c|}{0} &  \multicolumn{2}{c|}{0} & \multicolumn{2}{c|}{0} & \multicolumn{2}{c|}{1} \\ \hline\hline
BDSD & 3.40 & 3.03 & 2.40 & 2.14 & 3.90 & 3.41 & 0.9646 & 0.9713 \\  \hline
GSA & 3.96 & 3.83 & 2.80 & 2.71 & 3.73 & 3.49 & 0.9373 & 0.9373 \\ \hline \hline
AWLP & 2.85 & 2.41 & 1.98 & 1.69 & 2.16 & 1.97 & 0.9794 & 0.9822 \\  \hline
GLP & 3.42 & 2.79 & 2.39 & 1.97 & 2.51 & 2.24 & 0.9709 & 0.9768 \\ \hline \hline 
BAY & 2.62 & 2.49 & 1.85 & 1.75 & 2.67 & 2.37 & 0.9745 & 0.9789 \\  \hline
NLVD & 1.79 & 1.73 & 1.27 & 1.22 & 1.56 & 1.53 & 0.9877 & 0.9877 \\  \hline \hline
Avg. & 3.01 & 2.71 & 2.12 & 1.91 & 2.76 & 2.50 & 0.9691 & 0.9724 \\ \hline
\end{tabular}
}
\caption{Quantitative comparison on aerial images between fused products ({\it Fus}) and the associated restored images ({\it Rest}). In all cases, the quality metrics are improved by the proposed restoration method.}
\label{table_simulation_rgbnir}
\end{table}

\subsection{Evaluation on Aerial Images with Ground Truth}\label{sec_aerial}

These experiments were conducted on aerial images which were furnished to us by CNES. Four bands, blue (B), green (G), red (R) and near-infrared (I), were initially collected by matrix sensors (no push-broom) at a resolution of 10 cm by an aerial platform. Reference images at 30 cm were obtained by low-pass filtering and decimation in the Fourier domain with a MTF value of 0.15 cut frequency. These reference images, which are displayed in Figure \ref{fig_pelican}, allow an accurate quality assessment of the results. In this setting, we use the metrics RMSE, ERGAS, SAM and Q$2^n$ \cite{VivoneAlparoneChanussot2015}.

\begin{figure}[!t]
\footnotesize
\centering
\begin{tabular}{@{\hskip 0.03in}c@{\hskip 0.02in}c@{\hskip 0.01in}c@{\hskip 0.01in}c@{\hskip 0.01in}c}
\rotatebox{90}{\parbox[t]{0.8in}{\hspace*{\fill}Reference\hspace*{\fill}}} &
 \multicolumn{2}{c}{\includegraphics[trim= 28.9cm 28.7cm 8.3cm 8.5cm, clip=true, width=0.2\textwidth]{pelican1_gamma.png}} &
 \multicolumn{2}{c}{\includegraphics[trim= 28.9cm 28.7cm 8.3cm 8.5cm, clip=true, width=0.2\textwidth]{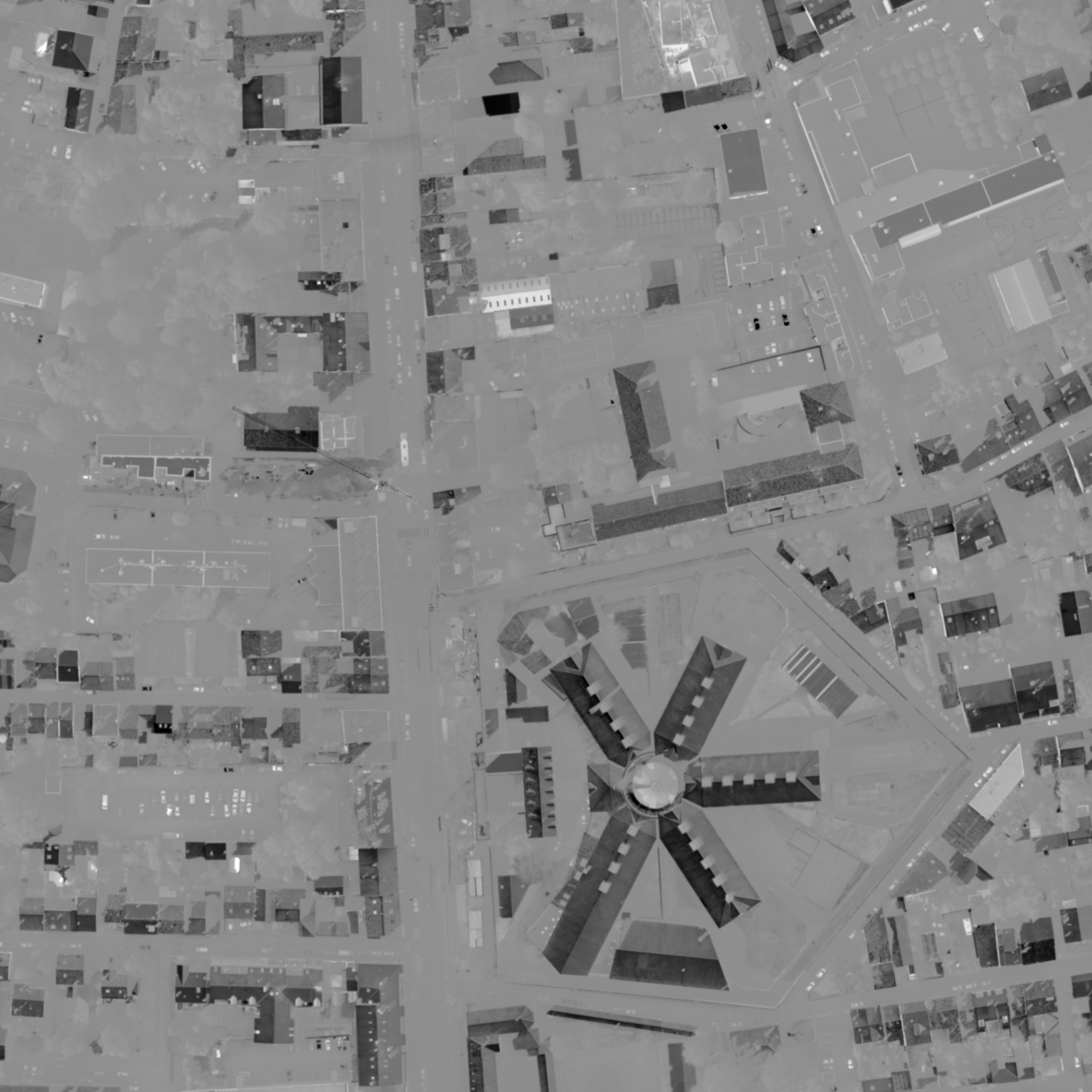}} \\
\rotatebox{90}{\parbox[t]{0.8in}{\hspace*{\fill}BDSD\hspace*{\fill}}} &
\includegraphics[trim= 28.9cm 28.7cm 8.3cm 8.5cm, clip=true, width=0.2\textwidth]{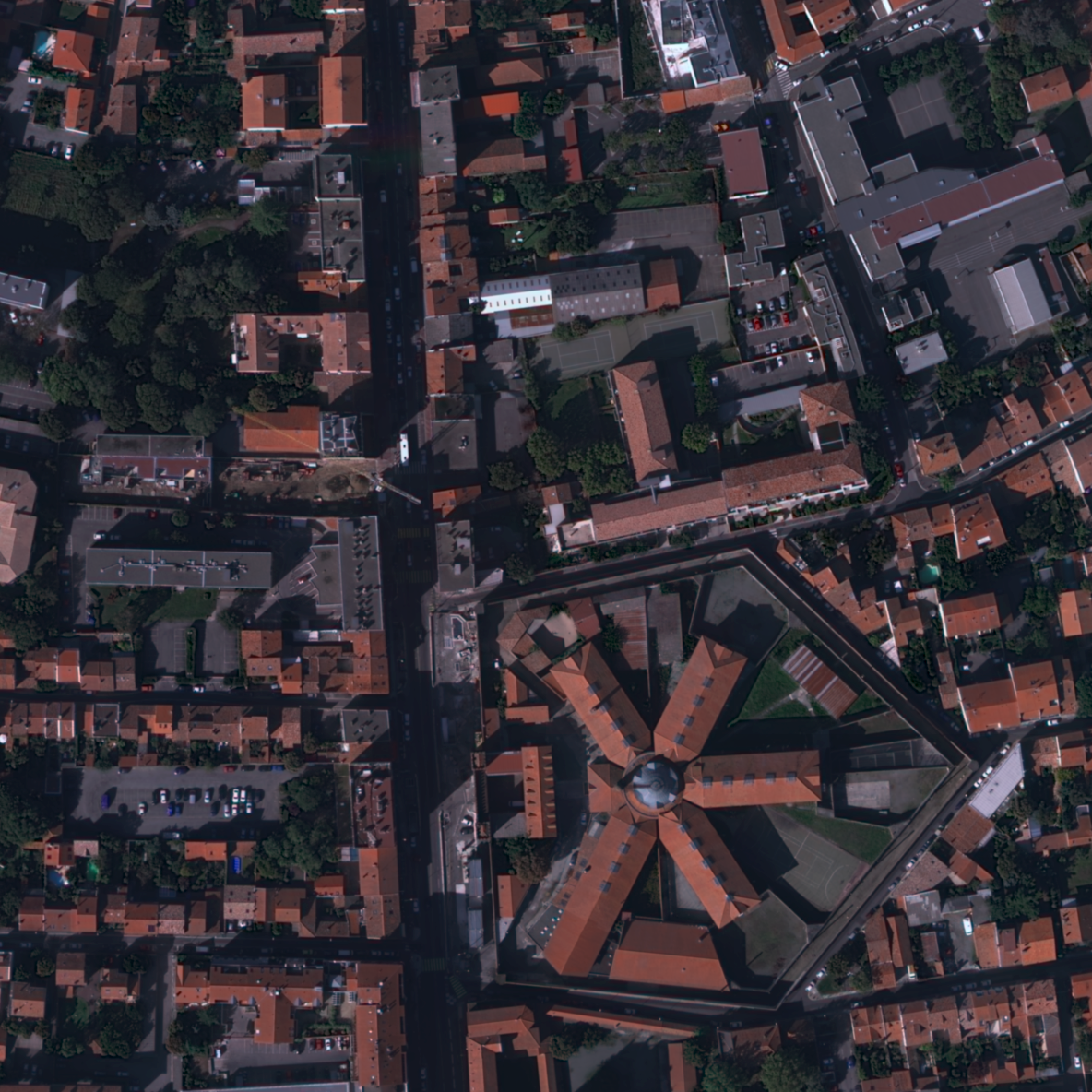} &
\includegraphics[trim= 28.9cm 28.7cm 8.3cm 8.5cm, clip=true, width=0.2\textwidth]{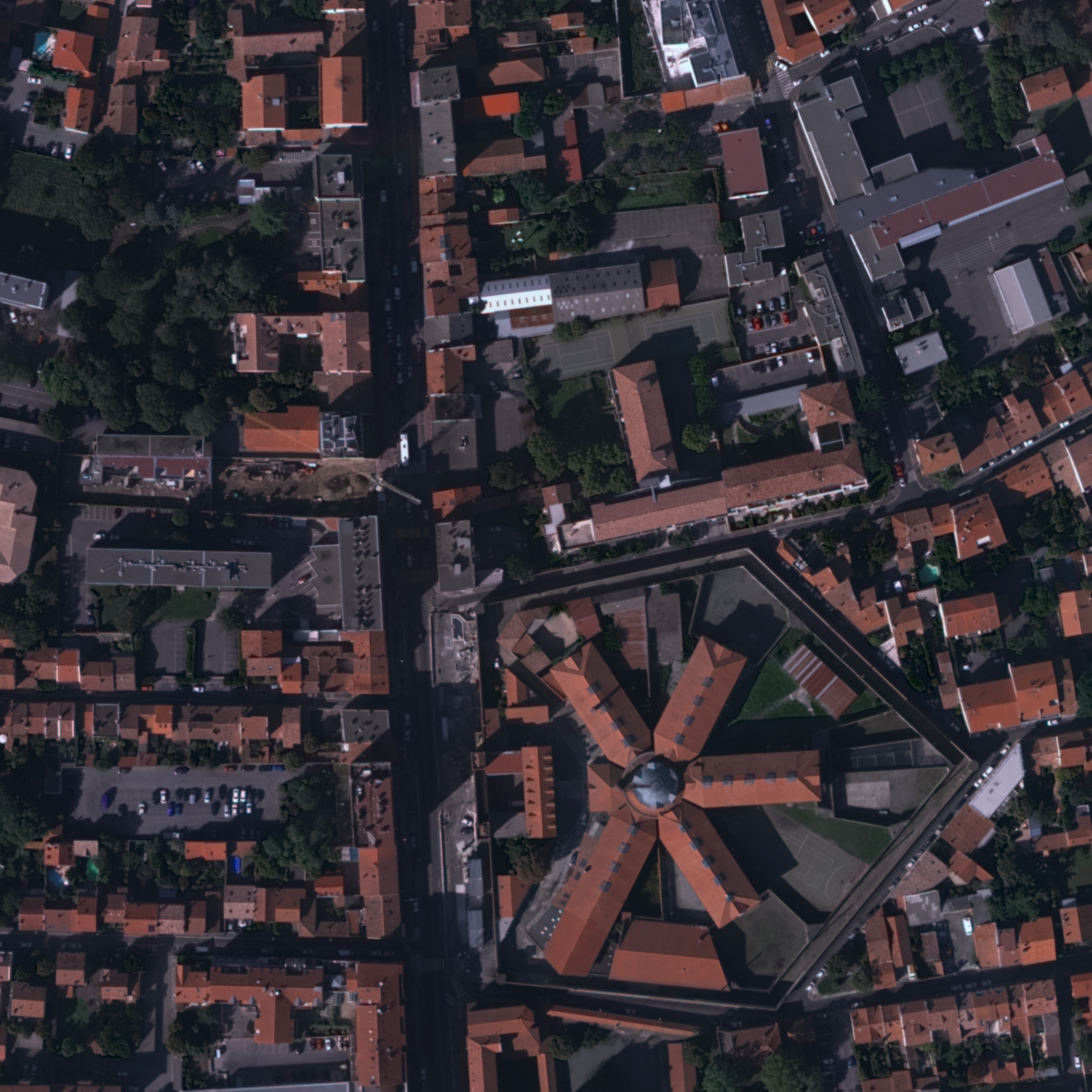} &
\includegraphics[trim= 28.9cm 28.7cm 8.3cm 8.5cm, clip=true, width=0.2\textwidth]{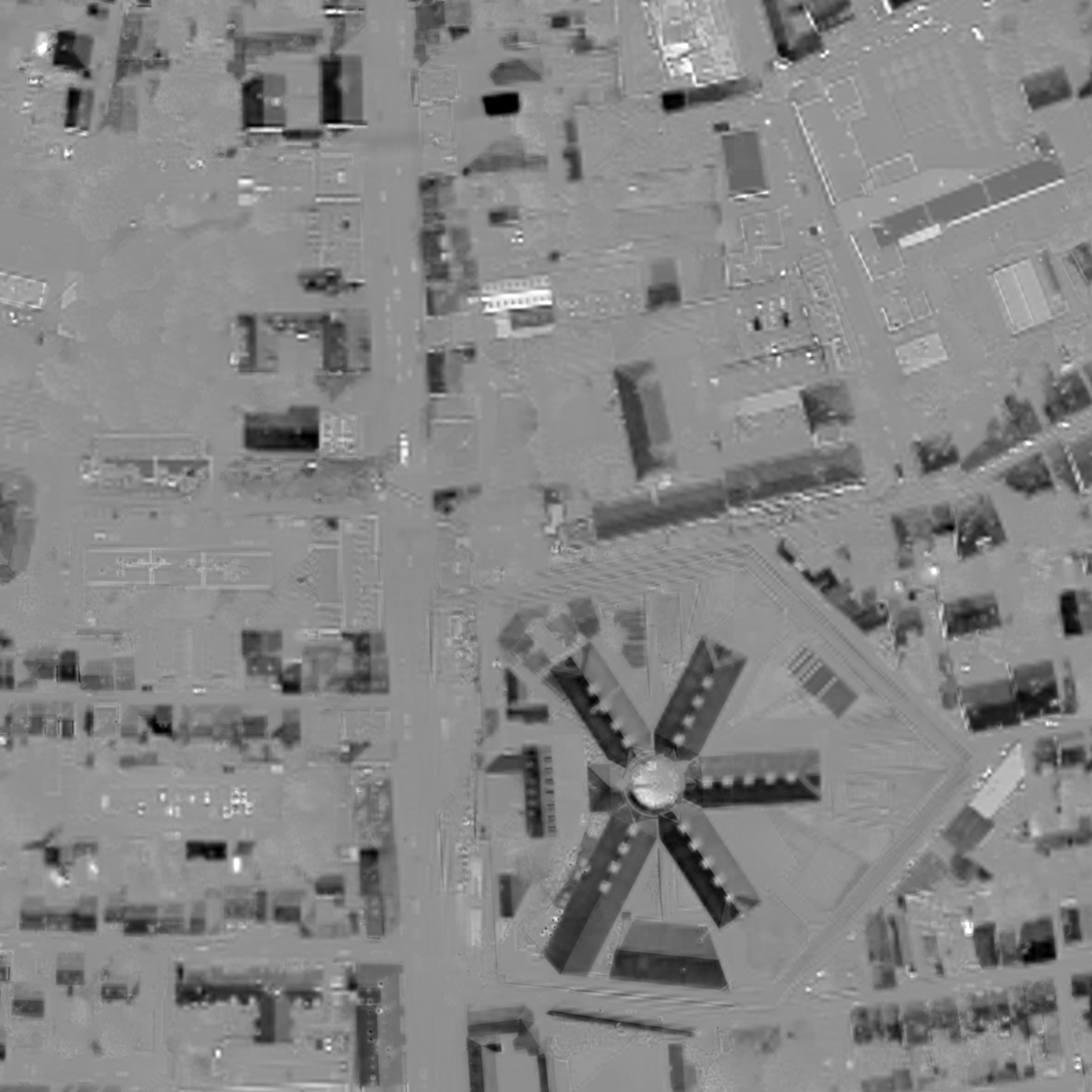} &
\includegraphics[trim= 28.9cm 28.7cm 8.3cm 8.5cm, clip=true, width=0.2\textwidth]{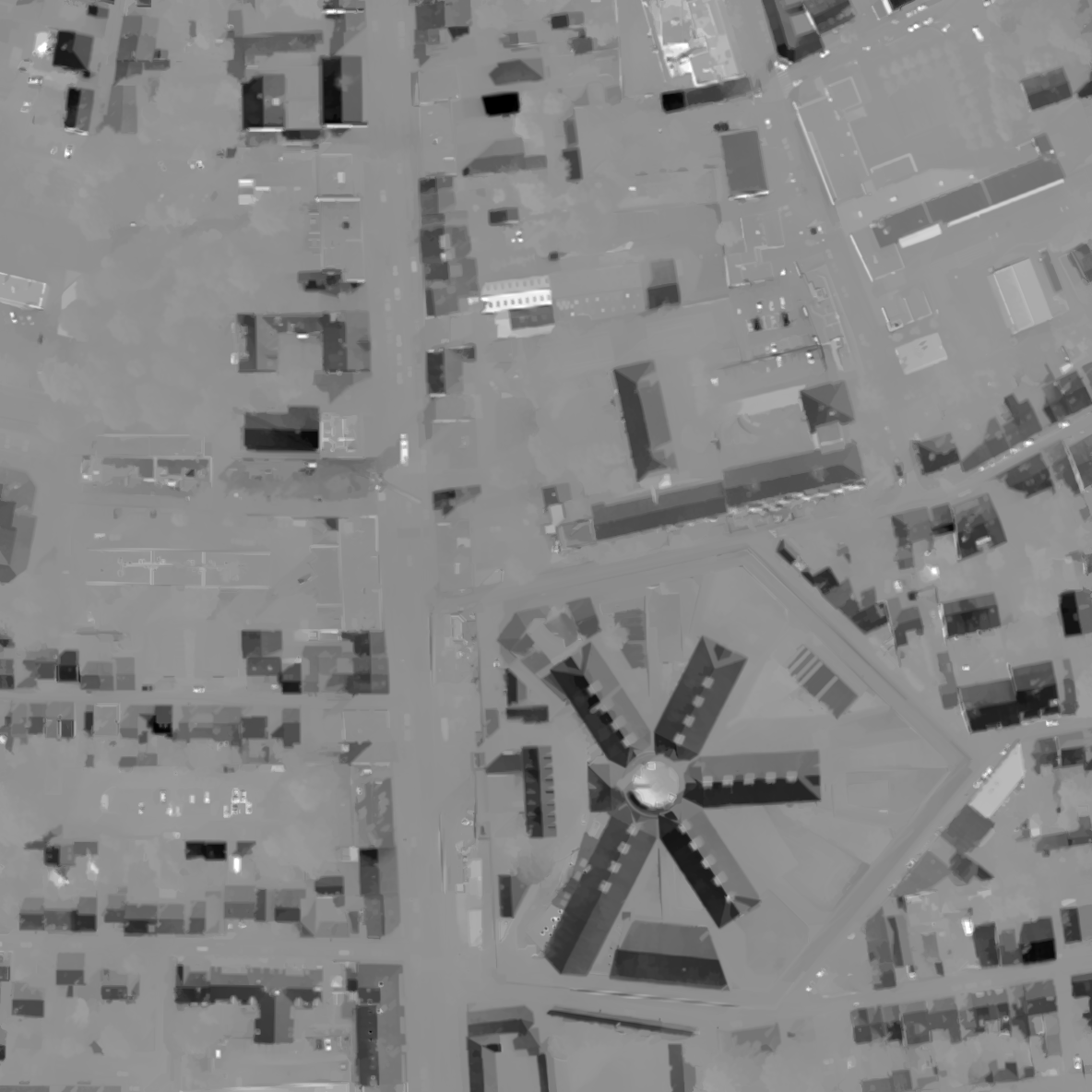} \\ 
\rotatebox{90}{\parbox[t]{0.8in}{\hspace*{\fill}AWLP\hspace*{\fill}}} &
\includegraphics[trim= 28.9cm 28.7cm 8.3cm 8.5cm, clip=true, width=0.2\textwidth]{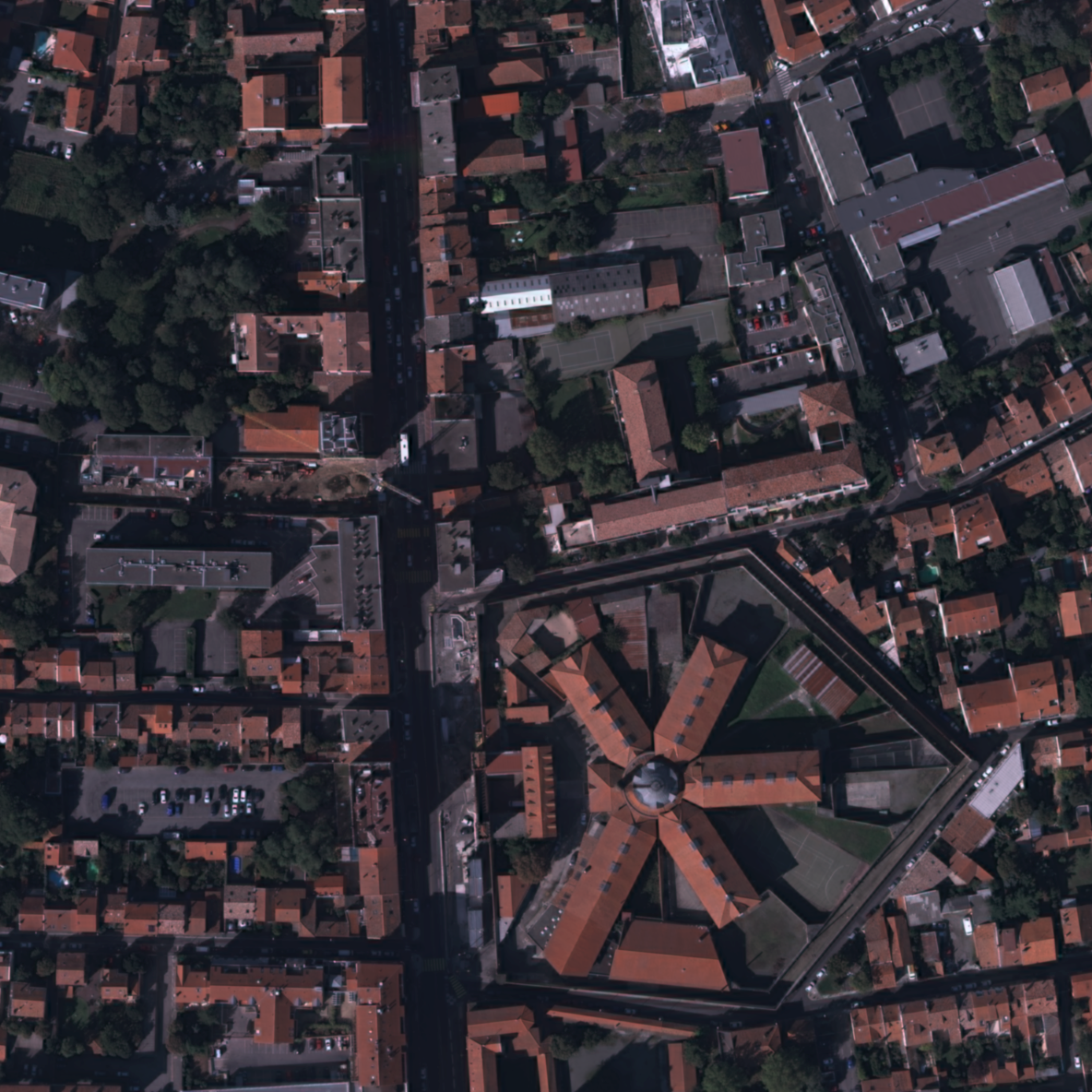} &
\includegraphics[trim= 28.9cm 28.7cm 8.3cm 8.5cm, clip=true, width=0.2\textwidth]{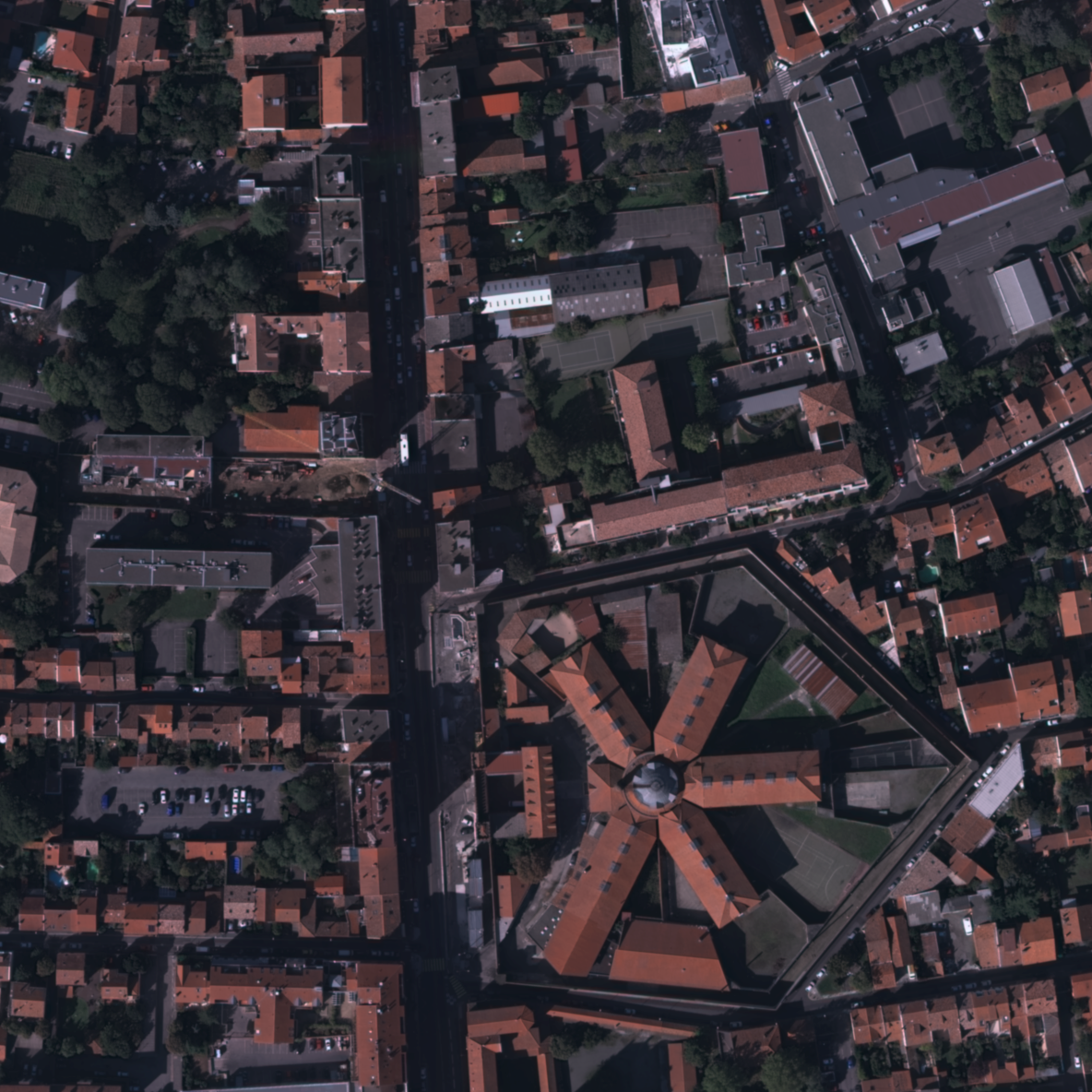} &
\includegraphics[trim= 28.9cm 28.7cm 8.3cm 8.5cm, clip=true, width=0.2\textwidth]{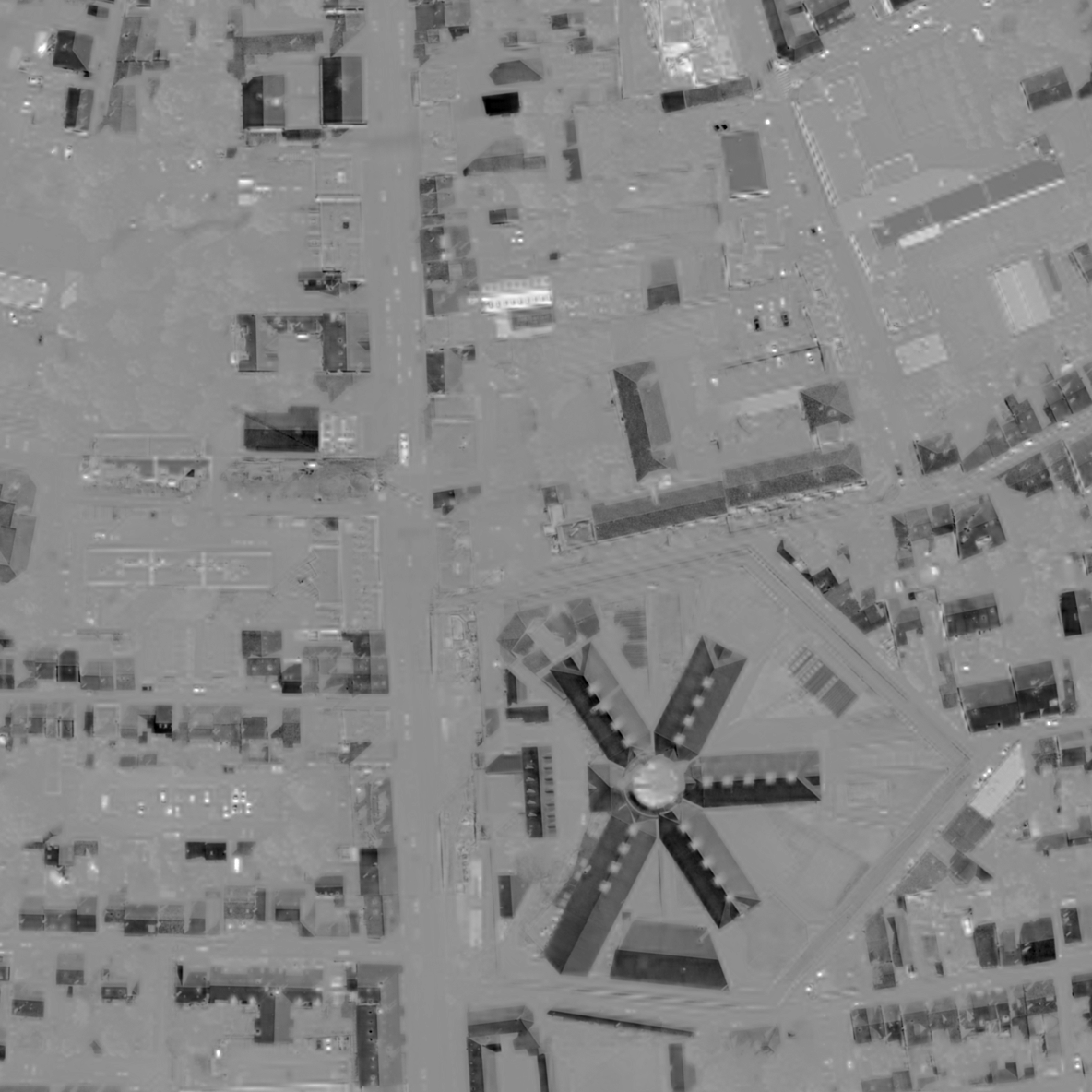} &
\includegraphics[trim= 28.9cm 28.7cm 8.3cm 8.5cm, clip=true, width=0.2\textwidth]{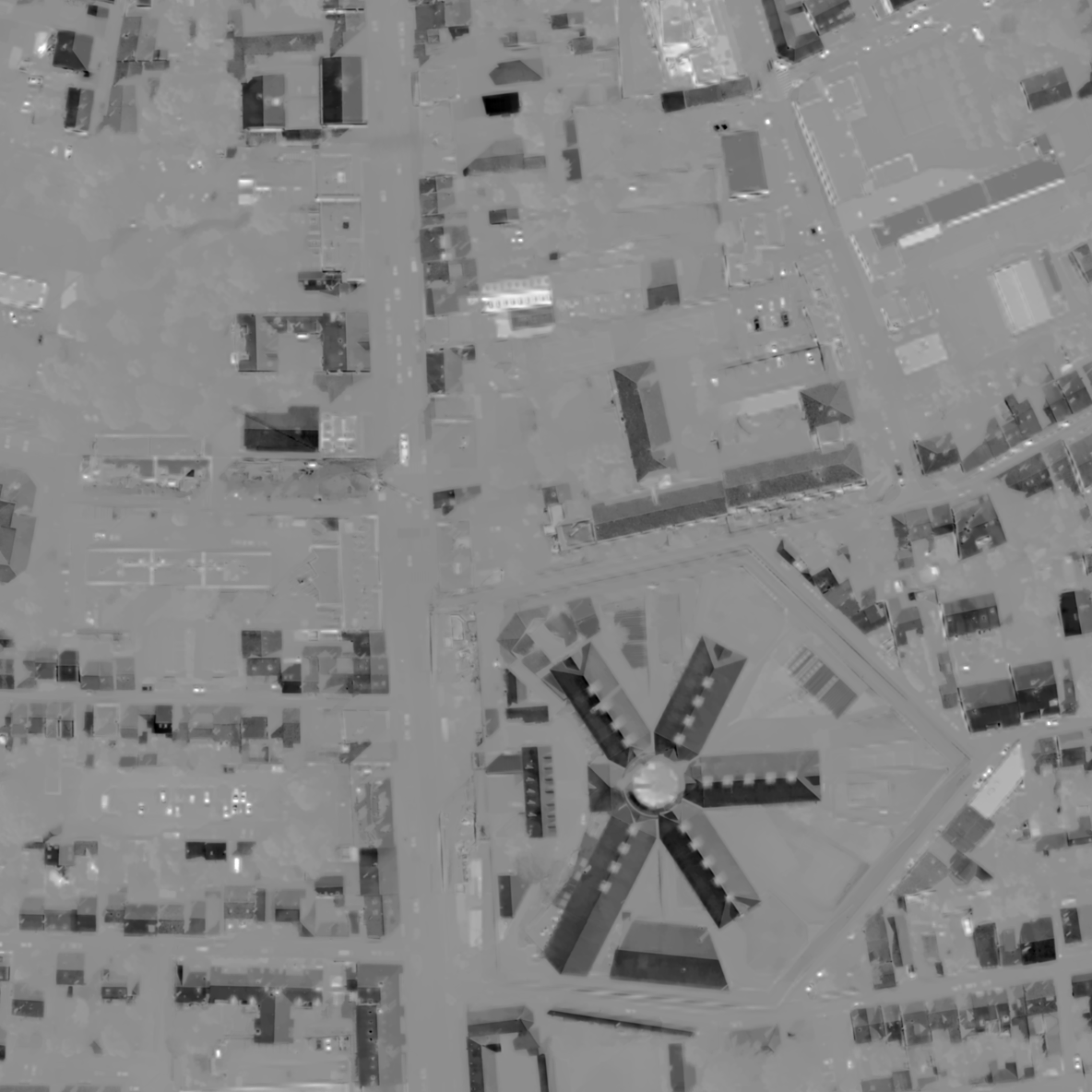} \\ 
\rotatebox{90}{\parbox[t]{0.8in}{\hspace*{\fill}BAY\hspace*{\fill}}} &
\includegraphics[trim= 28.9cm 28.7cm 8.3cm 8.5cm, clip=true, width=0.2\textwidth]{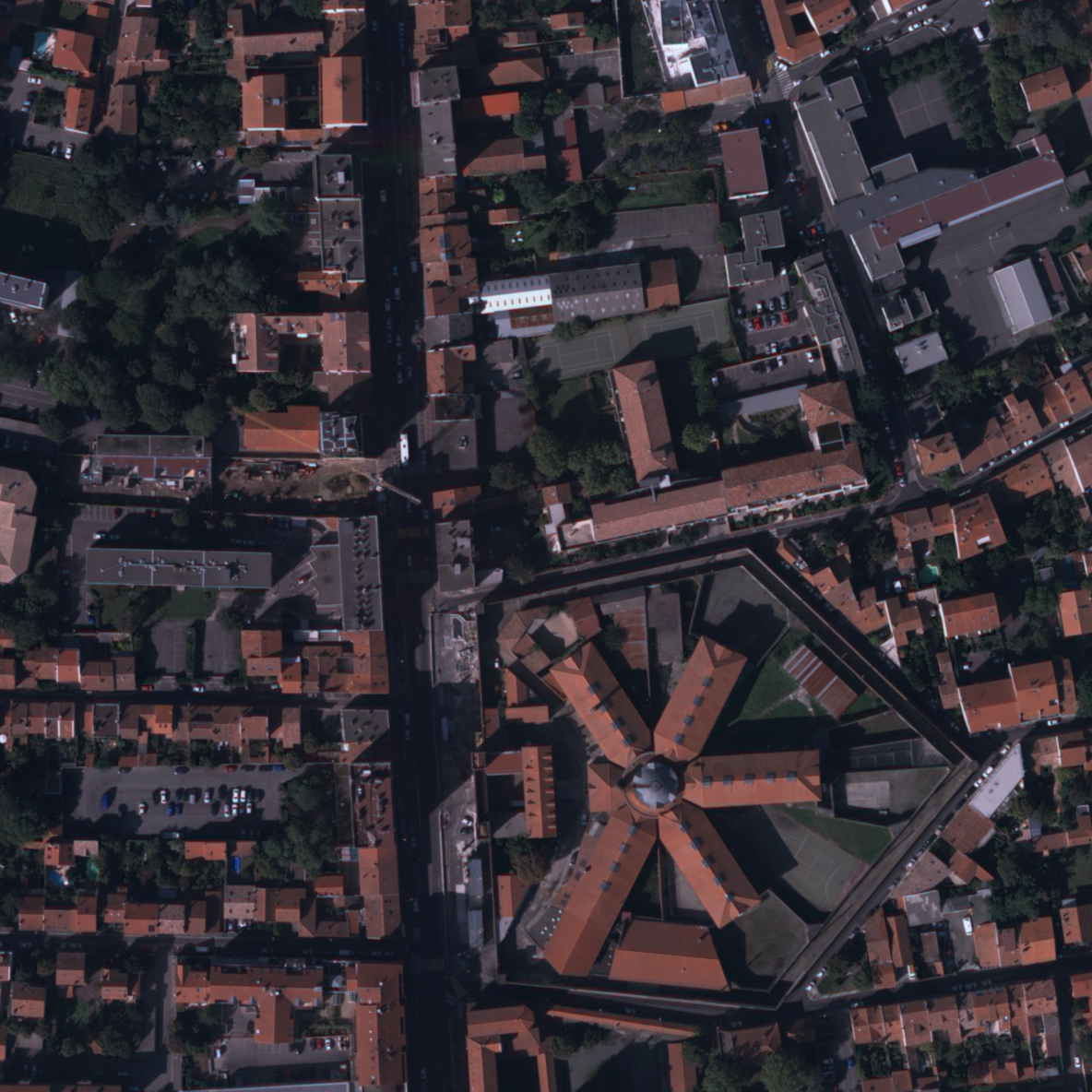} &
\includegraphics[trim= 28.9cm 28.7cm 8.3cm 8.5cm, clip=true, width=0.2\textwidth]{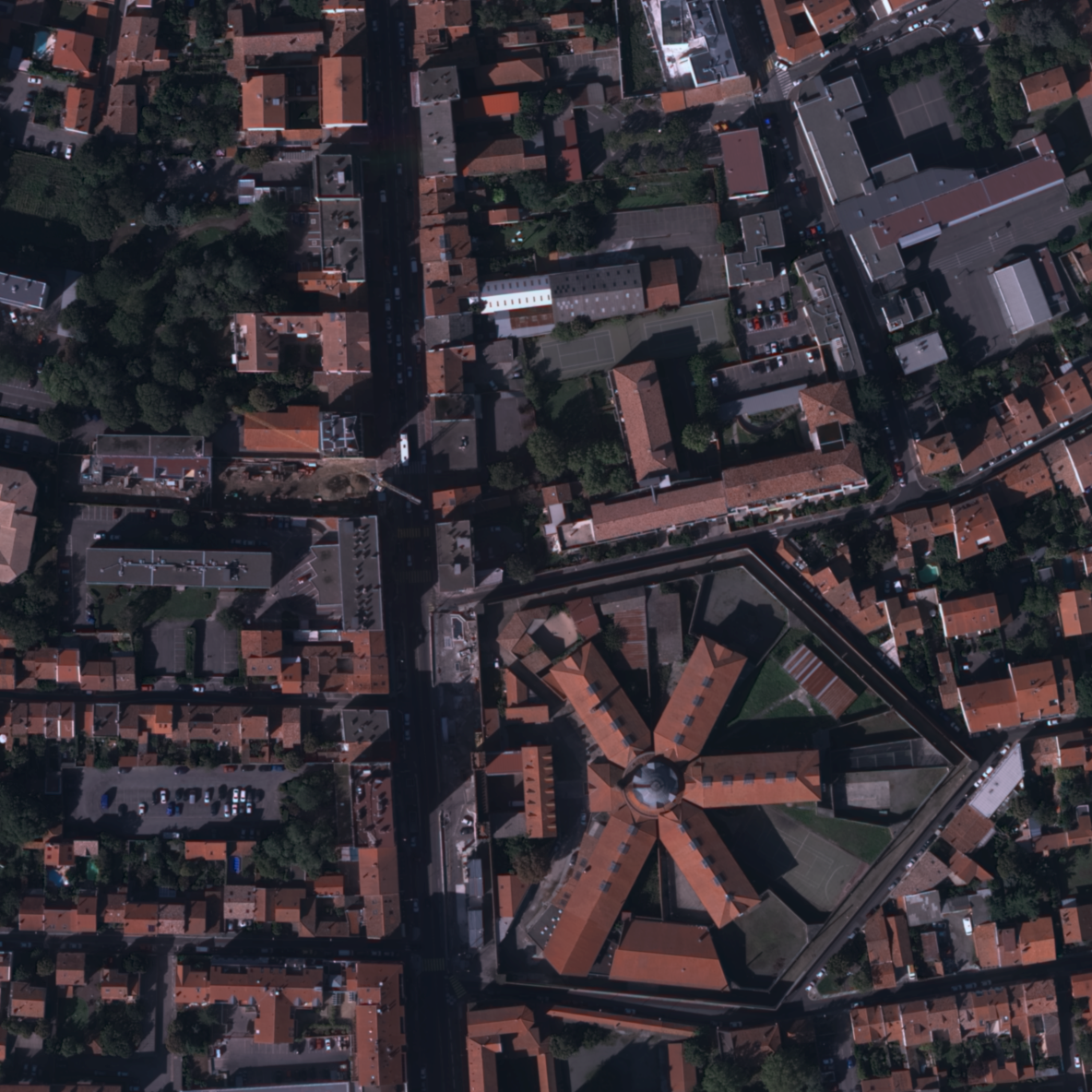} &
\includegraphics[trim= 28.9cm 28.7cm 8.3cm 8.5cm, clip=true, width=0.2\textwidth]{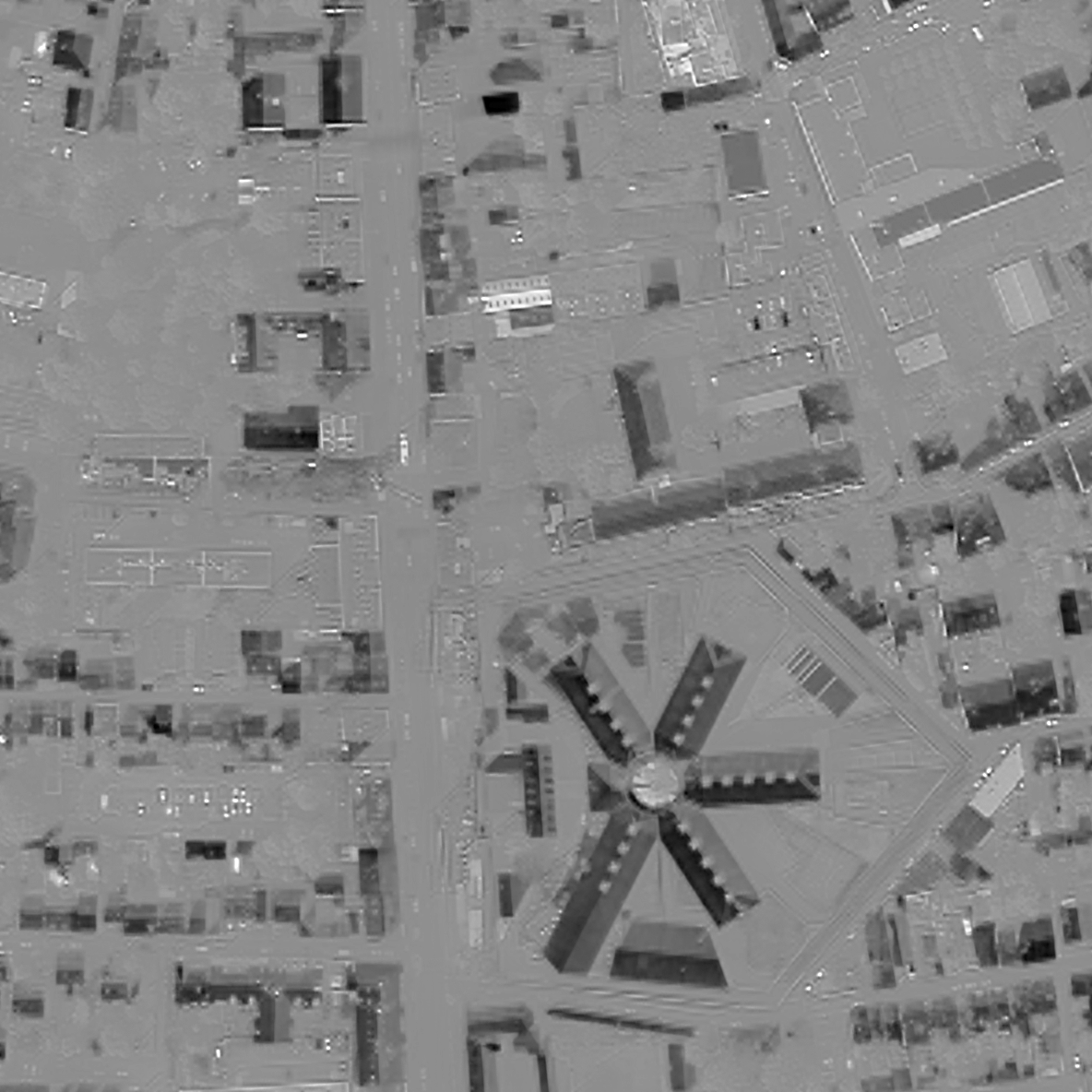} &
\includegraphics[trim= 28.9cm 28.7cm 8.3cm 8.5cm, clip=true, width=0.2\textwidth]{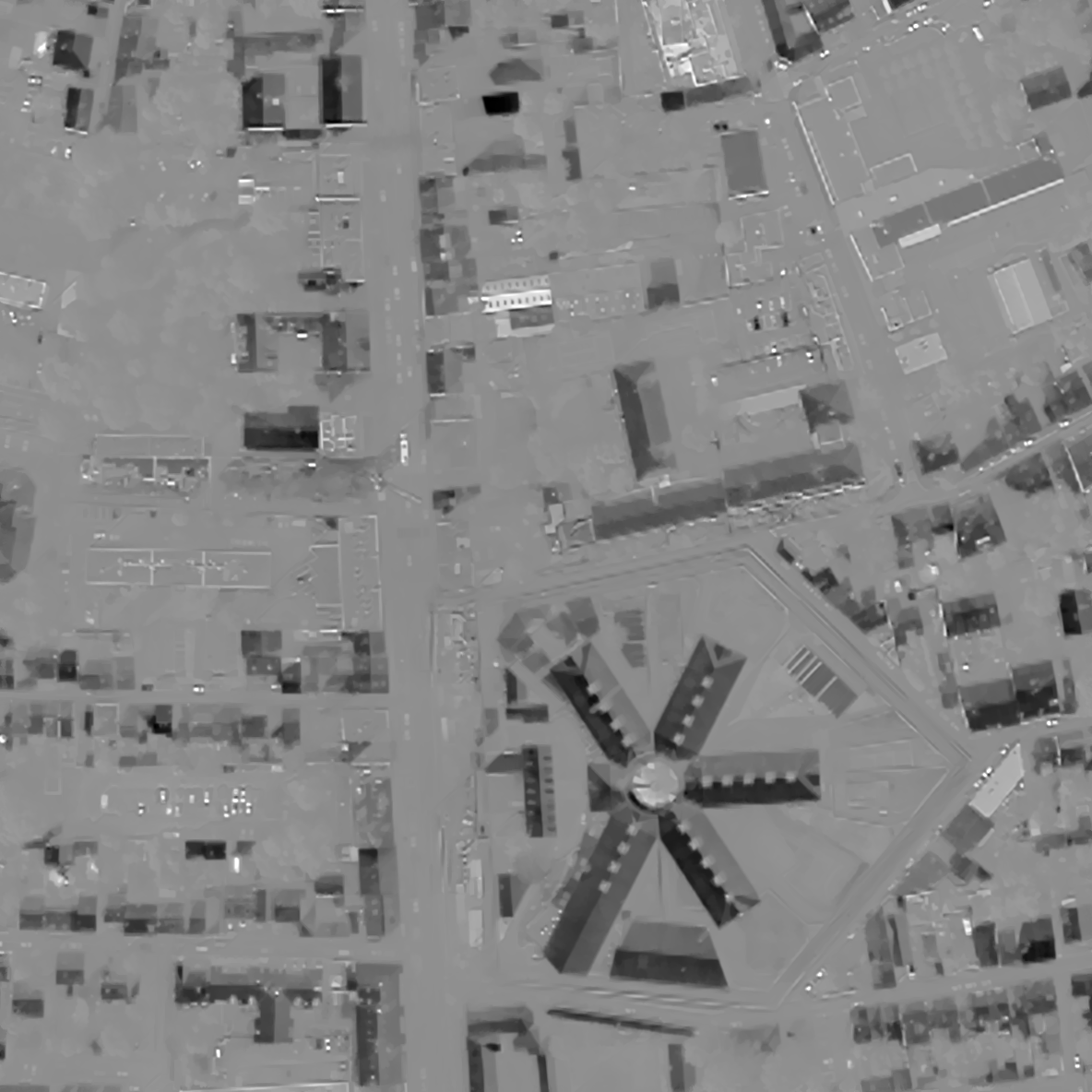} \\
 & Fused & Restored &  3rd PC fused & 3rd PC restored
\end{tabular}
\caption{Visual comparison on aerial images between RGB fused products and their associated restored images. The 3rd PCs are shown after linear rescaling of the displayed intensities. For visualization purposes, a gamma correction of factor $0.75$ has been applied to the RGB images. The proposed restoration chain mitigates aliasing, color spots and blurring effects.}
\label{fig_simulation_rgbnir}
\end{figure}

The PAN images at 30 cm were obtained by  averaging the original bands at 10 cm with coefficients $\alpha_B=0.1$, $\alpha_G=0.4$ and $\alpha_R=\alpha_I=0.25$, and then applying the same downsampling procedure. The selected values for the averaging coefficients are realistic and statistically acceptable on a variety of landscapes. On the other hand, the low-resolution MS channels have a spatial resolution four times lower than that of the PAN and were simulated from the original bands at 10 cm by low-pass filtering and decimation in the Fourier domain with a cutting frequency of $0.35$. 

We report in Table \ref{table_simulation_rgbnir} the quantitative results for each fused and restored product on the 4-band dataset. The proposed method improves the quality metrics in all cases. The restoration is particularly efficient in recovering the spatial quality (RMSE and ERGAS) in MRA fusion (AWLP and GLP) and the spectral quality (SAM) in CS fusion (BDSD and GSA), which is in line with previous analysis \cite{VivoneAlparoneChanussot2015}. The increase in Q2$^n$ reveals that our post-processing reduces luminance, contrast and spectral distortions. Although NLVD provides results of high quality, the metrics of the restored images being slightly better proves the beneficial impact of the restoration chain.

For visual quality assessment, Figure \ref{fig_simulation_rgbnir} displays close-ups of some pansharpened images, the restored results and their respective 3rd PCs. In the 3rd PCs, we observe that the aliasing patterns covering the scene are mitigated after restoration of all initially fused products. By looking at the RGB images, our method reduces color spots and blurring effects, thus improving both spectral and spatial qualities.

\subsection{Evaluation on Pl\'eiades Imagery}

We test the performance of our method on data acquired by Pl{\'e}iades, which produces PAN images at 70 cm and blue, green, red, and near-infrared bands at 2.8 m. The MTF for the PAN has a value of 0.15 at cut frequency, while this value is greater than 0.26 for the MS components resulting in strong aliasing. The size of the high-resolution images is $2048\times 2048$.

The lack of ground truth makes more difficult a quantitative evaluation. We use the {\it Quality with No Reference} (QNR) index proposed in \cite{Alparone2008}, which is a combination of spectral ($D_{\lambda}$) and spatial ($D_s$) distortions. Table \ref{table_pleiades} displays the quantitave results for each fused and restored product on a set of Pl\'eiades data, a crop of which is displayed in Figure \ref{fig_satellite_data}. The method decreases the spatial distorsion $D_s$, accounting for geometric features. Since we reduce aliasing in spectral bands, we increase inter-channel correlation, thus increasing $D_\lambda$. Overall, we succeed in improving the QNR of the fused images.

\begin{table}[!t]
\centering
\centering
{\small
\setlength\tabcolsep{2.6pt}
\renewcommand{\arraystretch}{1.05}
\begin{tabular}{|c|c|c|c|c|c|c|}
\hline
\multirow{2}{*}{Methods} & \multicolumn{2}{c|}{$\text{D}_{\lambda}$} & \multicolumn{2}{c|}{$\text{D}_s$} & \multicolumn{2}{c|}{QNR} \\ \cline{2-7}
 & Fus & Rest & Fus & Rest & Fus & Rest \\ \hline\hline
Reference & \multicolumn{2}{c|}{0} &  \multicolumn{2}{c|}{0} & \multicolumn{2}{c|}{1} \\ \hline\hline
BDSD & 0.0288 & 0.0305 & 0.0759 & 0.0654 & 0.8975 & 0.9061\\  \hline
GSA & 0.0411 & 0.0432 & 0.0812  & 0.0759 & 0.8810 & 0.8842  \\  \hline\hline
AWLP & 0.0482 & 0.0521 & 0.1026  & 0.0908  & 0.8541 &  0.8618 \\  \hline 
GLP & 0.0351 & 0.0397 & 0.0963  & 0.0865 & 0.8720  & 0.8772 \\  \hline\hline  
BAY &  0.0211 & 0.0225 & 0.0501 & 0.0439 & 0.9299 & 0.9346 \\  \hline
NLVD & 0.0206 & 0.0215  & 0.0417 & 0.0389 & 0.9386 & 0.9404 \\  \hline\hline
Avg. & 0.0325 & 0.0349 & 0.0746 & 0.0669 & 0.8955 & 0.9007  \\ \hline
\end{tabular}
}
\caption{Quantitative comparison on Pl{\'e}iades data between fused products ({\it Fus}) and the associated restored images ({\it Rest}). The QNR values improves after restoration. See text for details.}
\label{table_pleiades}
\end{table}

\begin{figure}[!p]
\footnotesize
\centering
\begin{tabular}{c@{\hskip 0.02in}c@{\hskip 0.01in}c@{\hskip 0.01in}c@{\hskip 0.01in}c}
\rotatebox{90}{\parbox[t]{1.1in}{\hspace*{\fill}BDSD\hspace*{\fill}}} &
\includegraphics[trim= 4.5cm 7.5cm 21.5cm 18.5cm, clip=true, width=0.2\textwidth]{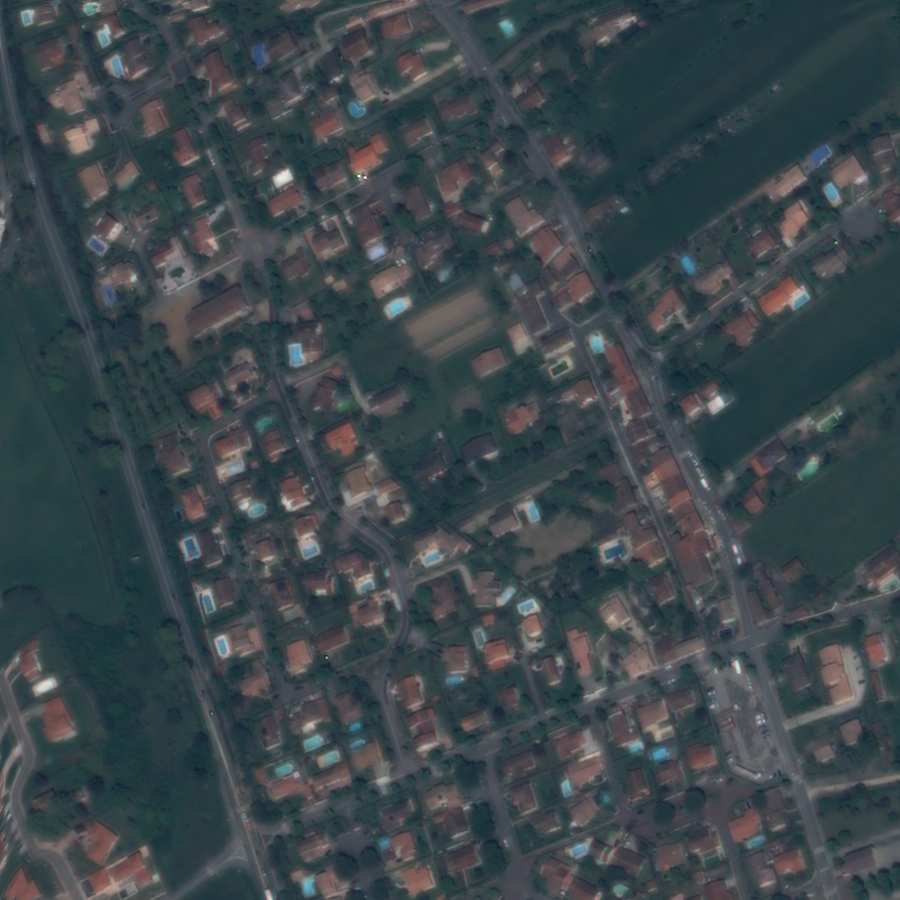} &
\includegraphics[trim= 4.5cm 7.5cm 21.5cm 18.5cm, clip=true, width=0.2\textwidth]{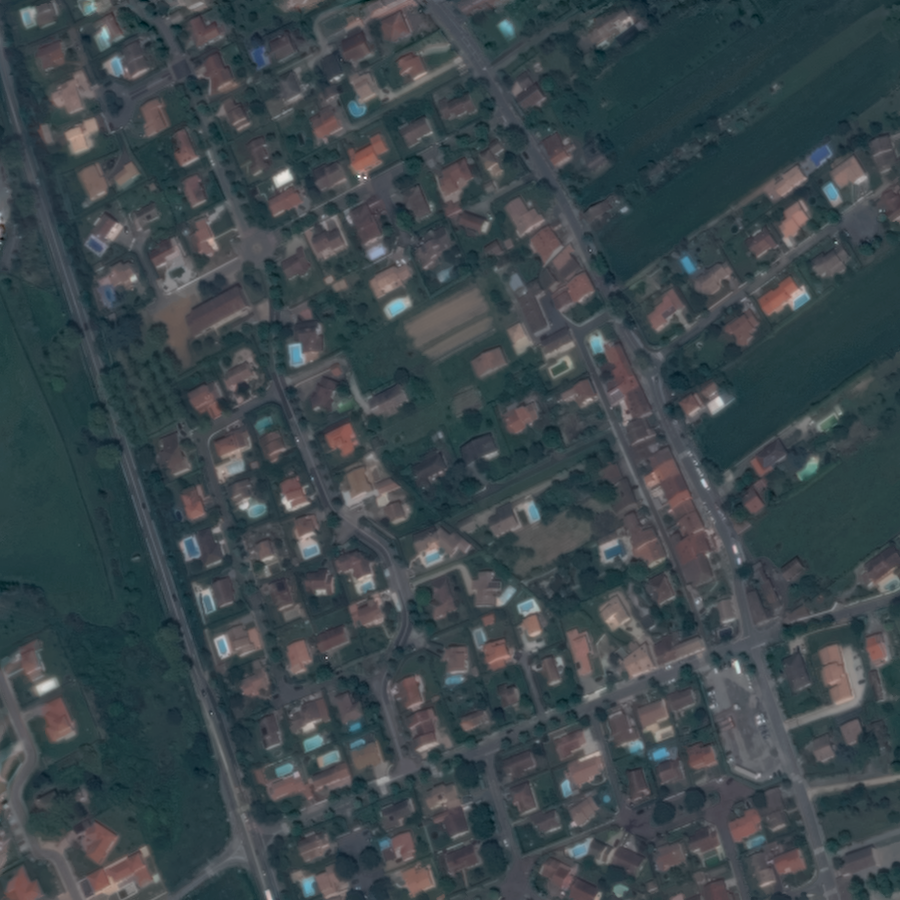} &
\includegraphics[trim= 4.5cm 7.5cm 21.5cm 18.5cm, clip=true, width=0.2\textwidth]{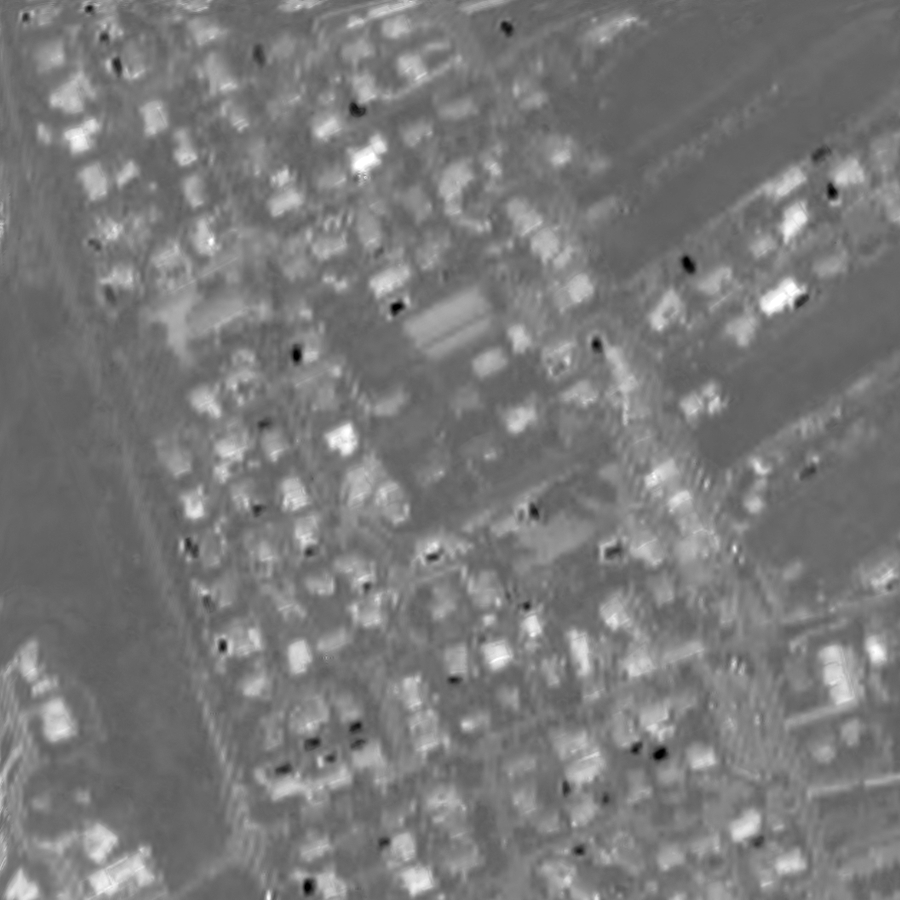} &
\includegraphics[trim= 4.5cm 7.5cm 21.5cm 18.5cm, clip=true, width=0.2\textwidth]{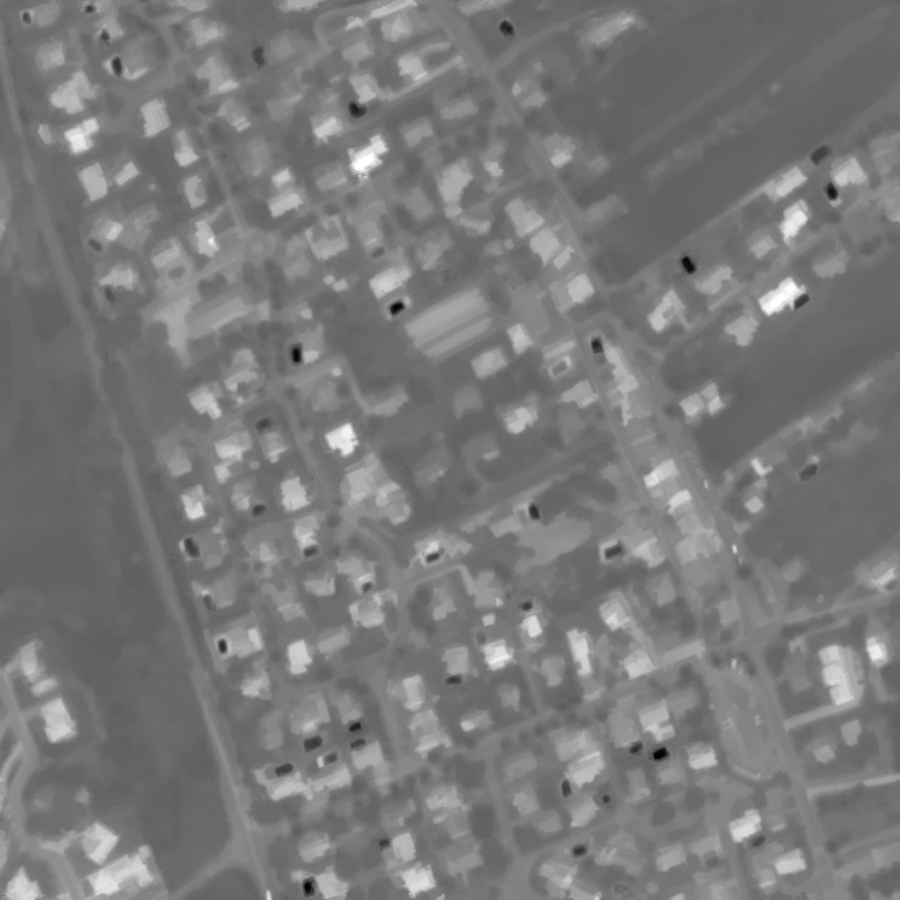} \\
\rotatebox{90}{\parbox[t]{1.1in}{\hspace*{\fill}GLP\hspace*{\fill}}} &
\includegraphics[trim= 4.5cm 7.5cm 21.5cm 18.5cm, clip=true, width=0.2\textwidth]{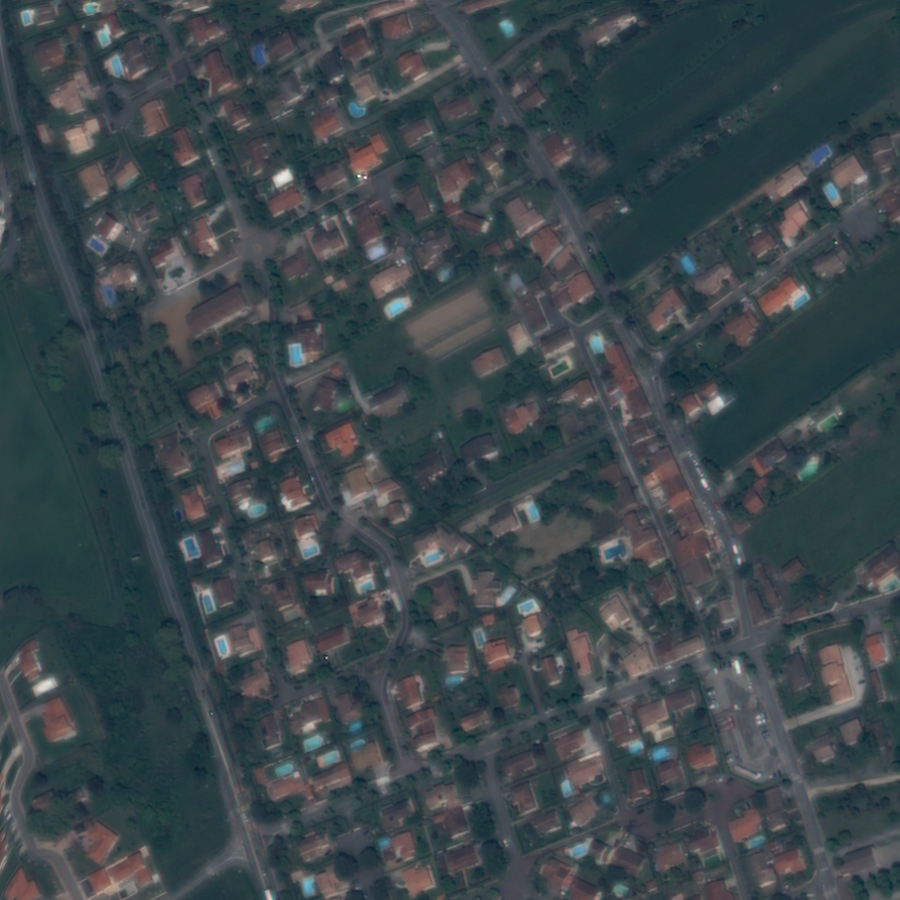} &
\includegraphics[trim= 4.5cm 7.5cm 21.5cm 18.5cm, clip=true, width=0.2\textwidth]{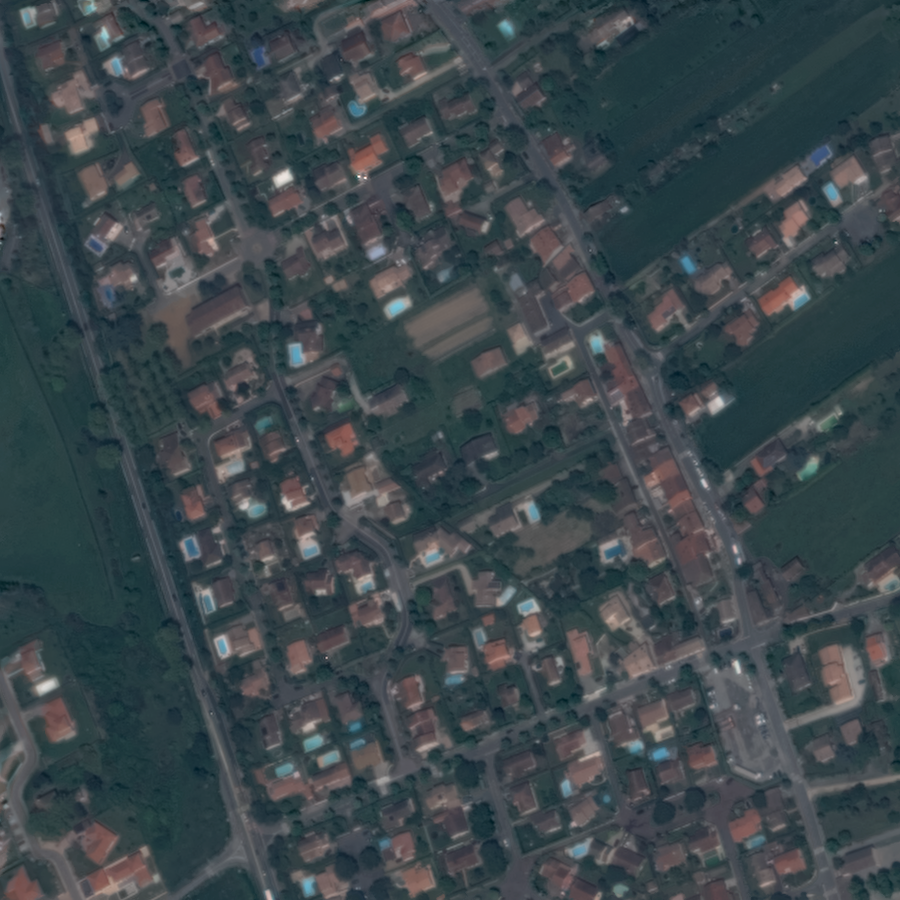} &
\includegraphics[trim= 4.5cm 7.5cm 21.5cm 18.5cm, clip=true, width=0.2\textwidth]{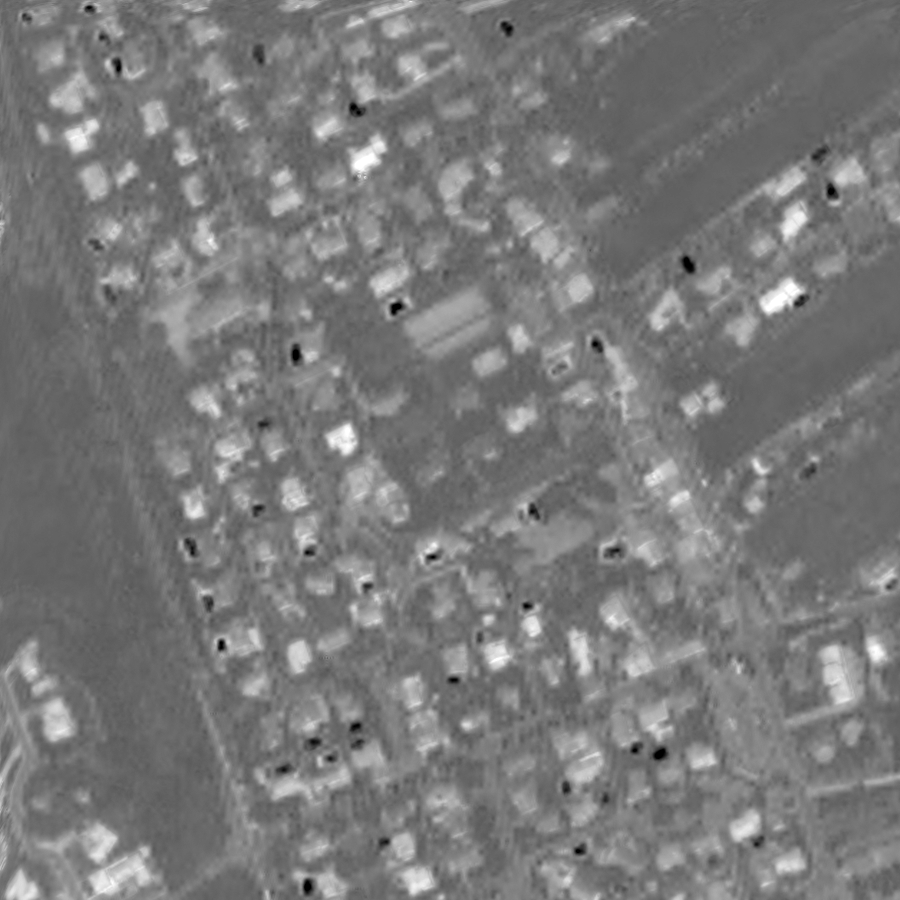} &
\includegraphics[trim= 4.5cm 7.5cm 21.5cm 18.5cm, clip=true, width=0.2\textwidth]{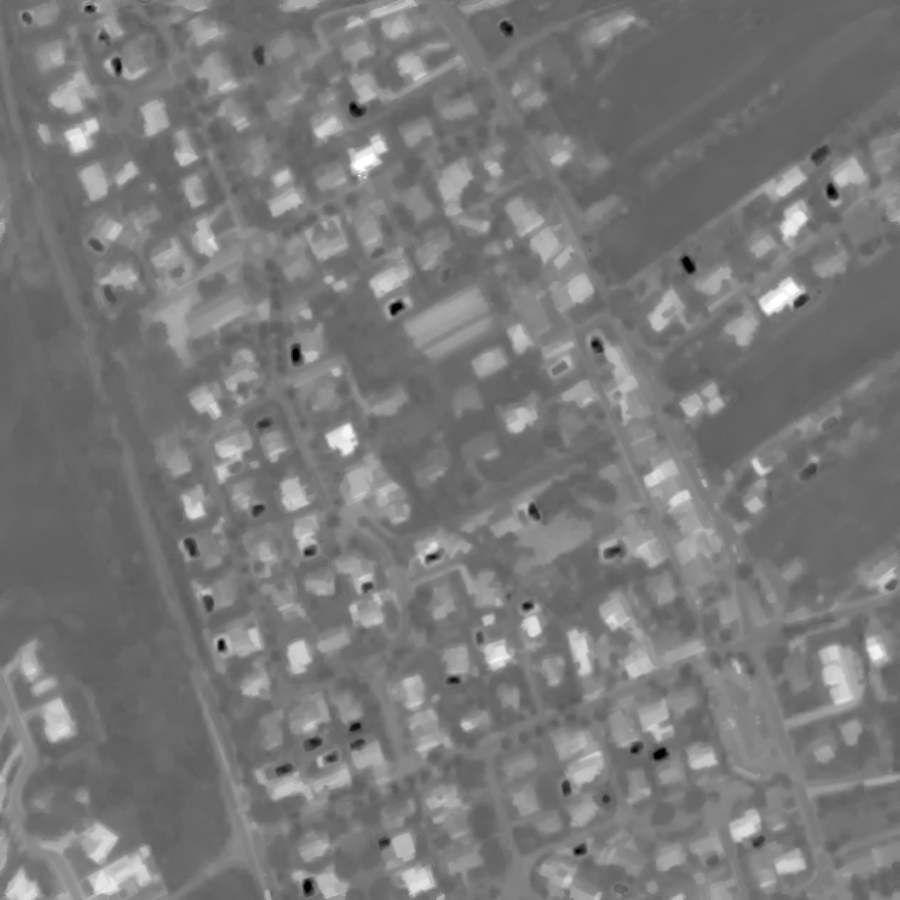} \\
\rotatebox{90}{\parbox[t]{1.1in}{\hspace*{\fill}NLVD\hspace*{\fill}}} &
\includegraphics[trim= 4.5cm 7.5cm 21.5cm 18.5cm, clip=true, width=0.2\textwidth]{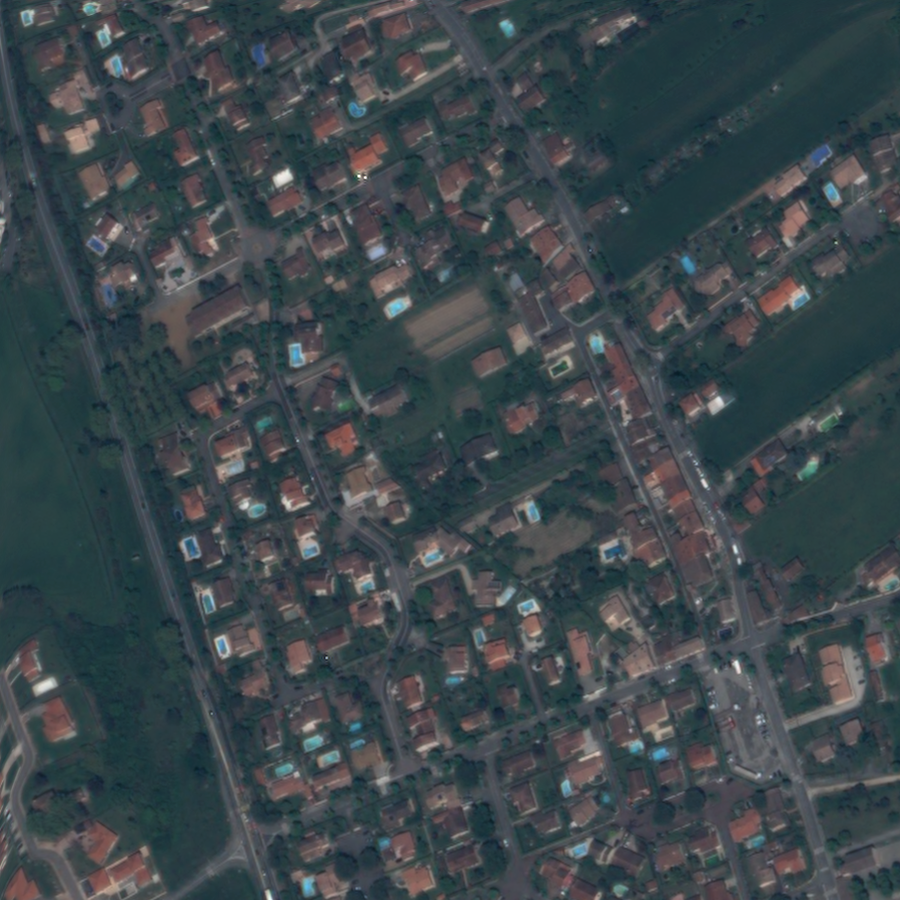} &
\includegraphics[trim= 4.5cm 7.5cm 21.5cm 18.5cm, clip=true, width=0.2\textwidth]{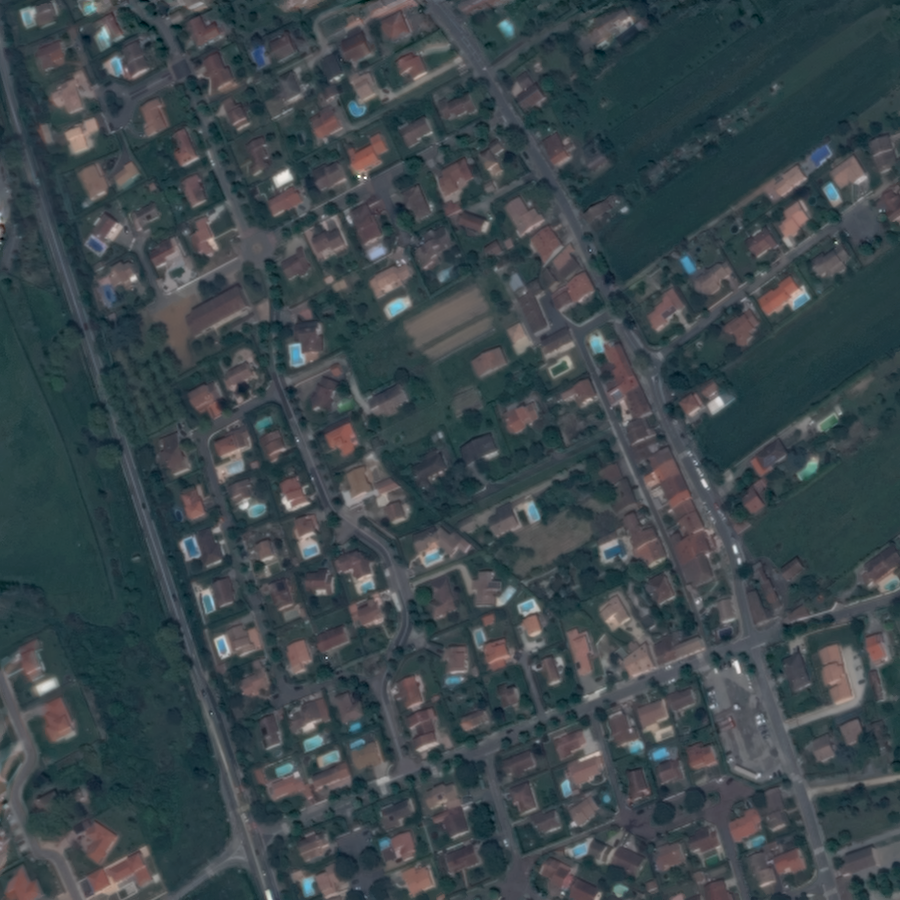} &
\includegraphics[trim= 4.5cm 7.5cm 21.5cm 18.5cm, clip=true, width=0.2\textwidth]{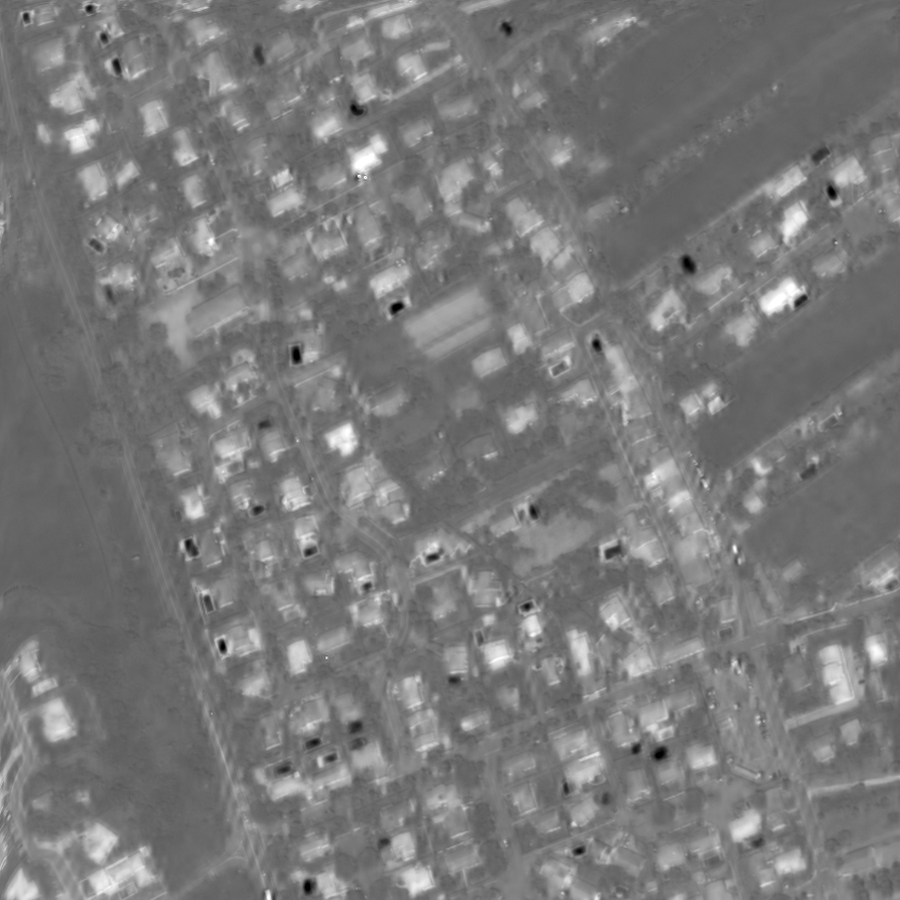} &
\includegraphics[trim= 4.5cm 7.5cm 21.5cm 18.5cm, clip=true, width=0.2\textwidth]{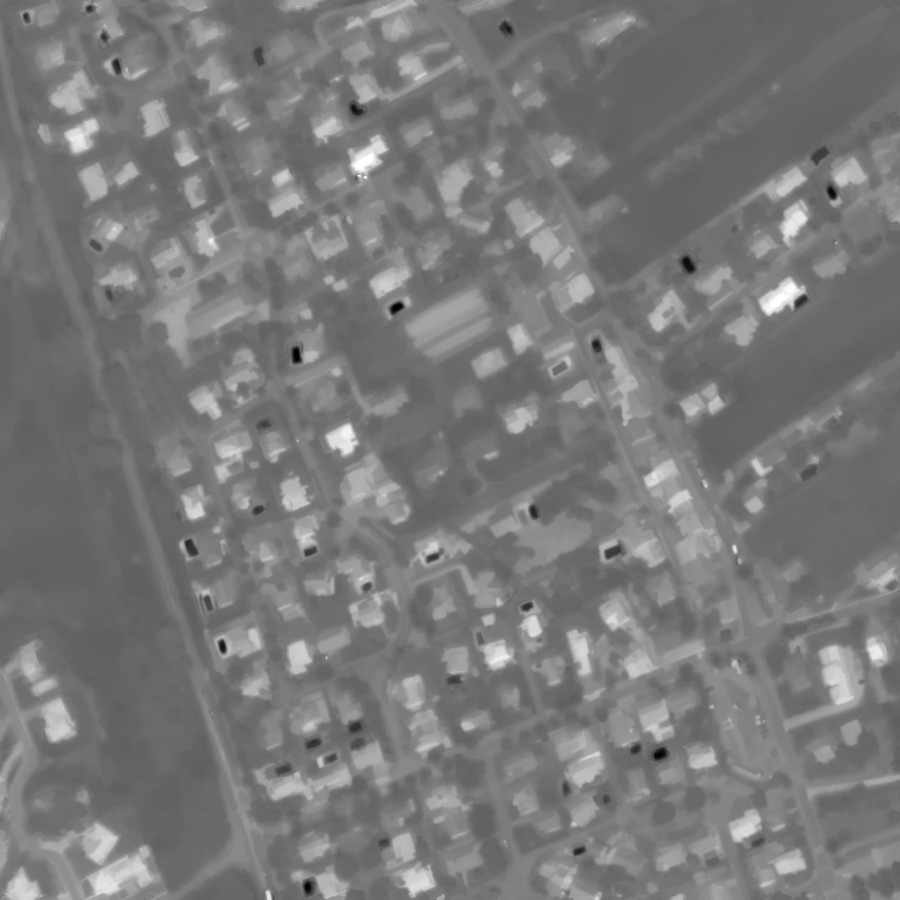} \\
\rotatebox{90}{\parbox[t]{1.1in}{\hspace*{\fill}BDSD\hspace*{\fill}}} &
\includegraphics[trim= 20cm 12cm 6cm 14cm, clip=true, width=0.2\textwidth]{pleiades_bdsd_gamma.png} &
\includegraphics[trim= 20cm 12cm 6cm 14cm, clip=true, width=0.2\textwidth]{pleiades_bdsd_rest_gamma.png} &
\includegraphics[trim= 20cm 12cm 6cm 14cm, clip=true, width=0.2\textwidth]{pleiades_bdsd_pc2.png} &
\includegraphics[trim= 20cm 12cm 6cm 14cm, clip=true, width=0.2\textwidth]{pleiades_bdsd_rest_pc2.png} \\
\rotatebox{90}{\parbox[t]{1.1in}{\hspace*{\fill}GLP\hspace*{\fill}}} &
\includegraphics[trim= 20cm 12cm 6cm 14cm, clip=true, width=0.2\textwidth]{pleiades_glp_gamma.png} &
\includegraphics[trim= 20cm 12cm 6cm 14cm, clip=true, width=0.2\textwidth]{pleiades_glp_rest_gamma.png} &
\includegraphics[trim= 20cm 12cm 6cm 14cm, clip=true, width=0.2\textwidth]{pleiades_glp_pc2.png} &
\includegraphics[trim= 20cm 12cm 6cm 14cm, clip=true, width=0.2\textwidth]{pleiades_glp_rest_pc2.png} \\
\rotatebox{90}{\parbox[t]{1.1in}{\hspace*{\fill}NLVD\hspace*{\fill}}} &
\includegraphics[trim= 20cm 12cm 6cm 14cm, clip=true, width=0.2\textwidth]{pleiades_nlvd_gamma.png} &
\includegraphics[trim= 20cm 12cm 6cm 14cm, clip=true, width=0.2\textwidth]{pleiades_nlvd_rest_gamma.png} &
\includegraphics[trim= 20cm 12cm 6cm 14cm, clip=true, width=0.2\textwidth]{pleiades_nlvd_pc2.png} &
\includegraphics[trim= 20cm 12cm 6cm 14cm, clip=true, width=0.2\textwidth]{pleiades_nlvd_rest_pc2.png} \\
& Fused & Restored & 3rd PC Fused & 3rd PC Restored
\end{tabular}
\caption{Visual comparison on Pl{\'e}iades data between RGB fused products and their associated restored images. The 3rd PCs are shown after linear rescaling of the displayed intensities. For visualization purposes, a gamma correction of factor $0.75$ has been applied to the RGB images. See text for details.}
\label{fig_pleiades_rgb}
\end{figure}

Figure \ref{fig_pleiades_rgb} shows close-ups of the pansharpened images, the restored results and their respective 3rd PCs. The initially fused products are affected by blur, see e.g.~road marks and vegetation, aliasing patterns, drooling effects surrounding objects such as swimming pools, and spectral spots. The proposed method is able to handle these issues and mitigate their negative impact on the visual assessment. In particular, the 3rd PCs illustrate how the conditional filtering eliminates most of the aliasing and spectral inaccuracies. Spatial details are further enhanced, especially at the edges of roads and buildings, and some texture from treetops and grass is recovered. In general terms, our approach increases spatial and spectral qualities regardless the fusion method previously used. 

\begin{figure}[t]
\footnotesize
\centering
\begin{tabular}{@{\hskip 0.01in}c@{\hskip 0.02in}c}
  \includegraphics[trim= 4.5cm 7.5cm 21.5cm 18.5cm, clip=true, width=0.35\columnwidth]{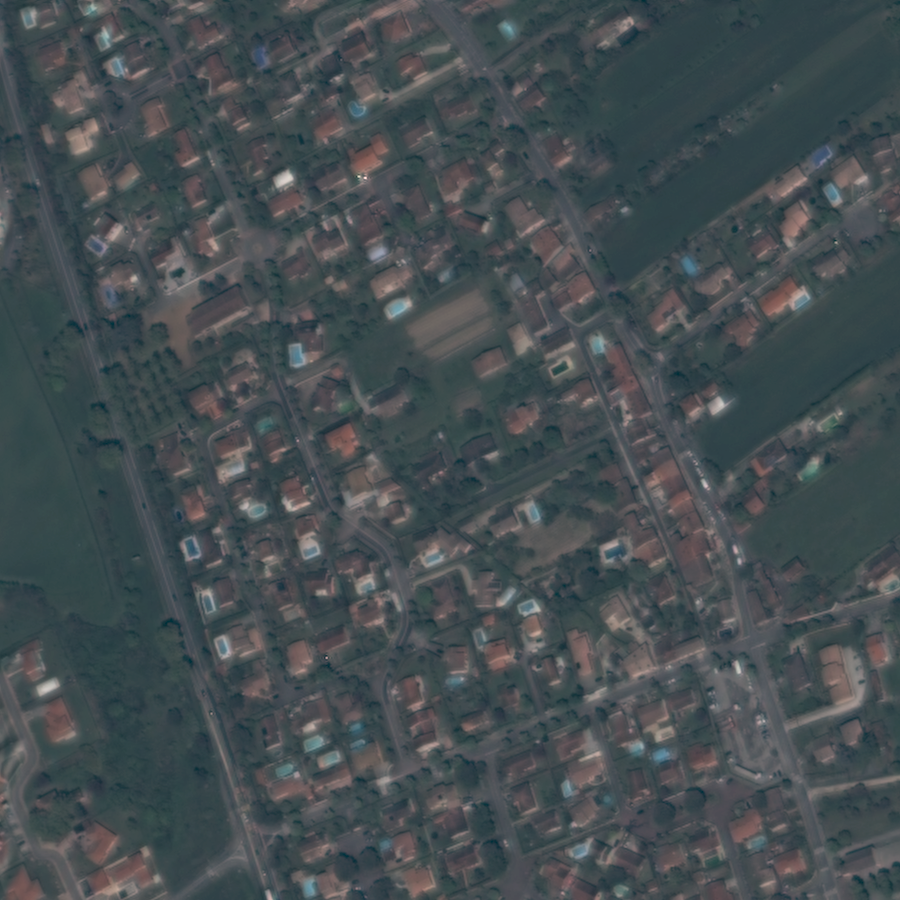} &
  \includegraphics[trim= 4.5cm 7.5cm 21.5cm 18.5cm, clip=true, width=0.35\columnwidth]{pleiades_glp_rest_gamma.png} \\
  \cite{LeeLee2010} & Ours
\end{tabular}
\caption{Visual comparison between our post-processing strategy and the one proposed in \cite{LeeLee2010}. We use as initialization the pansharpened image by GLP, which is shown in Figure \ref{fig_pleiades_rgb}. For visualization purposes, a gamma correction of factor 0.75 has been applied. Our approach better recovers spatial details as the contours of the buildings, the road marks and the texture from treetops and grass. On the contrary, blur and color spots remain in the restored image by \cite{LeeLee2010}.}
\label{fig_pleiades_restoration}
\end{figure}

Finally, we compare the performance of the proposed restoration method with the post-processing chain introduced in \cite{LeeLee2010}. For this purpose, we use the pansharpened image provided by GLP, which is shown in Figure \ref{fig_pleiades_rgb}.  Figure \ref{fig_pleiades_restoration} displays the restored images. Our technique better recovers spatial details as the contours of the buildings, the road marks and the texture from treetops and grass. On the contrary, blur and color spots remain in the restored image by \cite{LeeLee2010}. 

\section{Conclusion}\label{sec_conclusions}

We have introduced a restoration method to improve the quality of the pansharpened images independently of the fusion method.   An exhaustive performance evaluation have shown the ability of the proposed filtering to mitigate most of the blurring, spectral artifacts, aliased patterns and drooling effects.   This reveals that pansharpened images can be restored without any prior knowledge on the fusion strategy and of the formation model of the PAN and low-resolution MS bands.

\bibliographystyle{plain}
\bibliography{references}

\end{document}